\documentclass[11pt]{article}
\pdfoutput=1
\usepackage{graphics}
\usepackage{fullpage}
\usepackage{amsmath}
\usepackage{amsfonts}
\usepackage{amssymb}
\usepackage{mathrsfs}
\usepackage{dsfont}
\usepackage{algorithm}
\usepackage{algorithmic}
\usepackage{mathrsfs}
\usepackage[numbers]{natbib}

\usepackage{hyperref}

\usepackage{graphicx}

\usepackage[utf8]{inputenc}
\usepackage[english]{babel}
\newtheorem{theorem}{Theorem}
\newtheorem{repeattheorem}{Theorem}
\newtheorem{lemma}{Lemma}
\newtheorem{repeatlemma}{Restated Lemma}

\DeclareMathOperator*{\argmax}{arg\,max}

\newcommand{\proof}[1]{\textit{#1}}
\def\Halmos{$\square$}
\def\endproof{}

\def\E{\mathbb{E}}
\def\R{\mathbb{R}}

\def\Np{\mathbb{N}^+}
\def\P{\mathbb{P}}

\def\GP{\mathscr{GP}}

\def\One{\mathds{1}}
\def\st{\tilde{\sigma}}
\newcommand{\amax}[1]{\text{argmax}_{#1}}
\def\n1{^{n+1}}

\def\CRNKG{\text{KG}^{\text{CRN}}}
\def\PWKG{\text{KG}^{\text{PW}}}

\def\trans{^\intercal}

\def\tXn{\tilde{X}^n}
\def\ktb{k_{\bar\theta}}

\def\tb{\bar{\theta}}
\def\mub{\bar{\mu}}
\def\GP{\text{GP}}

\begin{document}

\title{Bayesian Optimization \\ Allowing for Common Random Numbers}

\author{Michael Pearce \\ Complexity Science \\ University of Warwick \\ Coventry, UK
\and
Matthias Poloczek \thanks{work was performed whilst at University of Arizona, Tuscon, AZ} \\ Uber AI \\ San Francisco, CA 
\and
Juergen Branke \\ Wawrwick Business School\\ Coventry, UK }

\maketitle

\newif\ifwithpics
\withpicstrue

\begin{abstract}
    Bayesian optimization is a powerful tool for expensive stochastic black-box optimization problems
such as simulation-based optimization or machine learning hyperparameter tuning. 
Many stochastic objective functions implicitly require a random number seed as input. By explicitly reusing
a seed a user can exploit common random numbers, comparing two or more inputs
under the same randomly generated scenario, such as a common customer
stream in a job shop problem, or the same random partition of training data into training and validation set for a machine learning algorithm.
With the aim of finding an input with the best average performance over infinitely many seeds, we propose a novel 
Gaussian process model that jointly models both the output for each seed and the average.
We then introduce the Knowledge Gradient for Common Random Numbers that iteratively determines a
combination of input and random seed to evaluate the objective and
automatically trades off reusing old seeds and querying new seeds, thus overcoming the
need to evaluate inputs in batches or measuring differences of pairs as suggested in 
previous methods. We investigate the Knowledge Gradient 
for Common Random Numbers both theoretically and empirically, finding it achieves
significant performance improvements with only moderate added computational cost.
\end{abstract}

\section{Introduction}
We consider the problem of expensive stochastic optimization with limited evaluations,
\begin{equation}
\argmax_{x\in X} \E[\theta(x,s)],
\end{equation}
where $\theta(x,s)$ is a real valued output, $X\subset \R^d$ is the solution space, usually given by box constraints for continuous variables, or a set of discrete alternatives. The parameter~$s$ represents all of the stochasticity in
the objective, i.e., $\theta(x,s)$ is deterministic. For example, $s$ may be the seed
of a pseudo random number generator that is called within a simulator. Hence evaluating
multiple $x$ with the same $s$ will reuse a set of common random numbers (CRN).
Alternatively, the seed $s$ and random number stream uniquely define a ``scenario" passed to
the objective function, and the aim of optimization
is to find an $x\in X$ that is the best averaged over all possible randomly generated
scenarios. Example applications include
\begin{itemize}
    \item[] {\bf Control and Reinforcement Learning:} $x$ are parameters of a control policy, $s$ defines a randomly
    generated environment (e.g. maze, race track, terrain) and $\theta(x,s)$ is final reward.
    \item[]{\bf Machine Learning:} $x$ are hyperparameters of a machine learning algorithm or model,
    $s$ defines a random split of training data into train and validation sets, and $\theta(x,s)$ is accuracy.
    \item[] {\bf Simulation Optimization:} In many optimization problems, a solution $x$ can only be evaluated by a stochastic simulator $\theta(x,s)$ whose seed~$s$ we may choose.
\end{itemize}
In this work we empirically investigate the following two simulation optimization applications.
\begin{itemize}
    \item[]{\bf Inventory Management:} $x$ are target inventory levels below which more stock is ordered,
    $s$ defines a random stream of customers and $\theta(x, s)$ is profit.
    \item[]{\bf Base Location:} $x$ are spatial locations of ambulance bases, $s$ defines 
    times and locations of patients randomly appearing across the map, and $\theta(x, s)$
    is average ambulance journey time.
\end{itemize}
From a surrogate modelling perspective, as a result of using CRN, the noise
corrupting the objective output has covariance for outputs with the same seed.
This is in contrast to the common assumption of independent noise for the objective outputs. For example, the seed $s$
may influence the difficulty of a randomly generated scenario, and the
performance of all solutions $x\in X$ degrades for difficult scenarios and improves
for easy scenarios. 

Traditionally, CRN has been exploited by considering the reduction in variance of performance 
differences, $\theta(x,s)-\theta(x',s)$, as CRN typically induces a positive correlation in noise, and
\begin{eqnarray*}
\text{Var}(\theta(x,\cdot)-\theta(x',\cdot))
=\text{Var}(\theta(x,\cdot))
+\text{Var}(\theta(x',\cdot))
-2\text{Cov}(\theta(x,\cdot),\theta(x',\cdot)).
\end{eqnarray*}
There have been several previous works that focus on evaluating pairs of candidates or multiple
comparisons either ``with CRN" or ``without CRN".

In this work we take a different perspective. The domain of the objective is the cross-product of
the \emph{solution space} and positive integer seeds $X\times \{1,2,....\}$ and we refer to
this domain as the \emph{acquisition space}.
Therefore, the surrogate model is defined over $X\times \Np$ and an optimization 
algorithm needs to propose input pairs $(x,s)\in X\times \Np$ and evaluate $\theta(x,s)$
to learn $\amax{x}\tb(x)=\argmax_x\E[\theta(x,\cdot)]$.
Given this  perspective, we  emphasize that the benefit in using CRN comes
from the \emph{emergent structure in the noise}, i.e., how the output for a single seed is uniquely
different from the average over seeds,
\begin{eqnarray}
\epsilon_s(x) = \theta(x,s)-\tb(x).
\end{eqnarray}
In particular, if $\epsilon_1(x)=o_1$ is the constant function, this implies that
$\amax{x}\tb(x)=\amax{x}\theta(x,1)$ and it is sufficient to optimize the single seed $s=1$.
Thus, first we propose a Gaussian process model for $\theta(x,s)$
that also yields a method for inferring $\tb(x)$ and is a generalization of standard models.
Second, we propose the Knowledge Gradient for Common Random Numbers ($\CRNKG$)
that quantifies the value of a new point in $X\times \Np$ for learning
the optimizer of the average over infinitely many seeds, $\amax{}\ \tb(x)$.
Optimizing $\CRNKG$ determines the most beneficial combination of
solution $x$ executed with seed $s$ to efficiently learn $\amax{x} \tb(x)$.
The $\CRNKG$ algorithm is therefore able to automatically trade-off the benefits of evaluating $x$ with a previously
evaluated seed, thereby utilizing CRN, and of evaluating $x$ with a fresh new seed, by simply maximizing the 
expected benefit. This removes both, the need to observe multiple $x$ simultaneously
in a batch with CRN or the need to consider differences in pairs of outputs evaluated with CRN.
However, we point out that our $\CRNKG$ algorithm can easily be extended to batch acquisition, e.g., using the technique of \cite{wu2016parallel}.
%

In the following section we briefly summarize related work, then formally define the problem in
Section~\ref{sec:probdef}. Section \ref{sec:method} describes and motivates the proposed
surrogate model and Section \ref{sec:acqfun} derives the new acquisition procedure
and discuses practicalities. In 
Section~\ref{sec:prev_methods} we draw parallels with a previous approach based on pairwise
sampling. An empirical evaluation on both synthetic experiments and
the two simulation optimization applications mentioned above are presented in Section~\ref{sec:numerical}.
The paper concludes in Section~\ref{sec:conclusion}.

\section{Literature Review} \label{sec:litrev}
The use of common random numbers (CRN) can be applied to any stochastic optimization problem where the user
can control the randomness of the objective. A typical use case in stochastic computer simulation
is Ranking and Selection, the problem of finding the best from a finite (small) set of uncorrelated solutions. In such a problem setting, a user is able to perform repeated evaluation of all solutions, see \cite{kim2013statistical}
and \cite{frazier2012tutorial} for a summary of frequentist and Bayesian techniques respectively. 
Combining CRN with ranking and selection has been considered with two-stage methods \citep{nelson1995using,chick2001new}
that initially sample all solutions multiple times to learn noise covariance structure and a second stage to exploit the
learnt structure. \cite{fu2004optimal} further investigate the second stage of the two stage process.
More recently, a sequential method has been proposed by \cite{gorder2019ranking} that keeps track of all
sampled seeds and uses the same series of seeds for all candidates.

When the candidate solutions have associated features that can inform simulation output, then surrogate models
can aid the optimization and enable search over much larger (possibly infinite) spaces $X$. Gaussian Random 
Fields allow to define a correlated prior over outputs that depends on similarity in inputs across
the space. Gaussian processes (GP) \citep{rasmussen2003gaussian}, or Kriging
\citep{ankenman2010stochastic}, are often employed when the search space is numerical, i.e., continuous or integer.
\cite{jones1998efficient} consider the optimization of a deterministic function using a Gaussian process.
\cite{huang2006global} and \cite{KGCP_scott2011correlated} among many others consider noisy functions 
assuming independent noise. For integer ordered spaces, or any lattice/network, one may employ 
Gaussian Markov Random Fields \citep{l2019gaussian} for faster computation.
The consequence of GP modelling with correlated noise has been considered by \cite{chen2012effects} when
assuming constant noise correlation across the solution space $X$. \cite{xie2016bayesian}
propose a method to combine a GP with CRN for optimization. They sample either a single solution or a pair under a new seed in each iteration.

In this work we consider the seed $s$ a (categorical) input to the objective $\theta(x,s)$ and the target of optimization
$\tb(x)$ is the objective with the $s$ argument ``integrated out". Hence this work is related to optimization of functions with (continuous) integrals \citep{toscano2018bayesian} or simulation optimization with an uncertain simulation input parameter
\citep{pearce2017bayesian}. Both methods sequentially determine a solution and input parameter 
in order to optimize the objective integrated over input parameters. In such a problem setting the surrogate model 
and data collection are defined over the multidimensional domain of decision variables and input parameters. 
However, in the CRN setting, the variable to be averaged out is categorical and there is no ``similarity"
over seeds. 
In this work we show how the structural assumptions of CRN lead to
a specific model design and interactions with the acquisition procedure. 
This results in a \textit{dynamic} acquisition search space yet the algorithm
still maintains minimal computational increase over an equivalent non-CRN algorithm.

\section{Problem Definition} \label{sec:probdef}
Let $\theta:X\times \Np \to \R$ be an expensive-to-evaluate, real valued function with arguments composed of a real valued solution $x\in X\subset \R^d$ and a nominal positive integer seed $s\in \Np$ and the domain is the \emph{acquisition space} $\tilde X = X\times \Np$. We refer to $\theta(x,s)$ as the \textit{objective function}. The random seed $s$ controls all stochasticity in the function, i.e., $\theta(x,s)$ is deterministic. 
The aim is to identify the solution $x$ from the \emph{solution space} $X$ that maximizes the expectation of the objective over random number streams
$$\argmax_x \bar\theta(x) = \argmax_x\E[\theta(x,\cdot)]$$
and we refer to $\tb(x)$ as the \textit{target}.
There is a limited budget of $N$ objective function calls, and for each call, the user can choose a seed $s$ and a decision variable $x$, then observe $y=\theta(x,s)$. 
Function evaluations may be collected sequentially so that after $n$ measurements the user may determine the $x$ and $s$ for the $(n+1)^{th}$ function evaluation. 

If every call to the function uses a new unique random seed, the problem reduces to standard stochastic optimization and the user only needs to determine $x$ values for each evaluation of $\theta(x,s)$. The problem considered here is therefore a more general setting that allows the reuse of random number seeds by making the argument $s$ explicit.



\section{A Surrogate Model for Simulation with Common Random \\ Numbers} \label{sec:method}

Given a budget of $N$ calls to $\theta(x,s)$, the proposed Bayesian optimization algorithm has two phases, an initialization phase where we evaluate a small number of candidates $n_{init}\ll N$, chosen as a space filling design in $X\times \{1,2,3,4,5\}$.
That is, we instantiate five (randomly chosen) seeds to collect data points that are then used to fit a Gaussian process model. The GP model is 
combined with an acquisition function (infill criterion) to sequentially
allocate the remaining $N-n_{init}$ points of the budget, updating the model after each new point and determining the next point.
We first describe our model for $\theta(x,s)$ and then propose the Knowledge Gradient
for Common Random Numbers in Section~\ref{sec:acqfun}.


\subsection{The Gaussian Process Generative Model}\label{sec:model}
A generative model is a probability distribution over all observable and unobservable quantities 
and such a model can be sampled to generate realizations of all variables thereby synthesizing data. 
Inference is the task of estimating the unobserved variables that are consistent with the generative 
model and the observed quantities. 
In the case of optimization with CRN, we desire a generative model with two properties. 
First, sampling outputs from the 
generative model assuming each output comes from a different seed must recover a model used without 
CRN. Second, the seeds are labeled with arbitrary numbers, in particular, there is  no exploitable ``neighborhood" between seeds.

Following previous works without CRN, we first assume that the target, $\tb(x)$, 
is a realization of a Gaussian process with constant prior mean $\bar \mu$ and covariance given by a kernel 
such as a $\frac{5}{2}$-Mat\'ern or squared exponential,
\begin{equation}
\tb (x) \sim \GP\big(\, \bar\mu, \,\, k_{\tb}(x,x') \,\big).
\end{equation}
When all seeds are unique, e.g., $s^i=i$, output $y$ values are generated by adding independent and identically distributed Gaussian noise 
$y\sim N(\tb(x), \sigma^2_\epsilon(x))$. Given $n$ solutions  $X^n=(x^1,...,x^n)$, the vector 
of outputs, $Y^n=(\theta(x^1,1),...,\theta(x^n,n))$, is assumed to be a single multivariate Gaussian 
random vector with the same and a covariance matrix composed of a kernel matrix and diagonal noise matrix
\begin{equation} \label{eq:genYiid}
Y^n \sim  N\big(\, \bar\mu, \,\, k_{\tb}(X^n,X^n) + \text{diag}(\sigma_\epsilon^2(X^n))\, \big).
\end{equation}
For $\theta(x,s)$ in the CRN setting, we require a kernel over $\tilde X = X\times \Np$ 
that when evaluated for unique seeds recovers the above covariance matrix. To
satisfy all zero off-diagonal elements for unequal seeds, we require a Kronecker
delta function over seeds (white noise), to model covariance
in outputs for the same seed we require another kernel over $X\times X$. 
We propose the following model for the objective,
\begin{equation}
\theta(x,s)\sim \GP\big(\, \bar\mu,\,\,  k_{\tb}(x,x') + \delta_{s's}k_\epsilon(x,x') \big),
\end{equation}
where $k_\epsilon(x,x')$ is the \emph{difference kernel} of the \emph{difference function} $\epsilon_s(x)$ between the target and the objective function for a particular seed
and must satisfy $k_\epsilon(x,x) = \sigma^2_\epsilon(x)$. 
We return to design of $k_\epsilon(x,x')$ shortly.
 $\mu^0(x,s)=\bar\mu$ is the constant prior mean.
Given a tuple of input pairs $\tilde X^n = ((x,s)^1, ...,(x,s)^n)$,
the generative distribution of $Y^n$ is thus
\begin{equation} \label{eq:genYcrn}
Y^n \sim  N\big( \, \bar\mu, \,\, k_{\tb}(X^n, X^n) + \One_{{S^n}} \circ k_{\epsilon}(X^n, X^n) \, \big),
\end{equation}
where $\circ$ denotes matrix element-wise (Hadamard) product and $\One_{{S^n}}\in [0,1]^{n\times n}$ 
is a binary masking matrix with elements equal to one at $(i,j)$ when $s^i=s^j$. 
Hence for the noise matrix, $\One_{{S^n}} \circ k_{\epsilon}(X^n, X^n)$, 
the diagonal and also any off-diagonal pairs where $s^i=s^j$ are
non-zero with corresponding covariance $k_\epsilon(x^i,x^j)$. 
The model encodes the functional form of the objective
as target and \emph{difference functions}, $\epsilon_s(x)$,
\begin{eqnarray}
\theta(x,s) &=& \tb(x) + \epsilon_s(x)
\end{eqnarray}
where the  $\epsilon_s(x)$ are independent
and identically distributed GP realizations
\begin{eqnarray}
\epsilon_1(x), \epsilon_2(x),...\sim \GP\big(\, 0, \,\, k_\epsilon(x,x') \,\big).
\end{eqnarray}
This model structure has multiple desirable properties. 
Firstly, by design it mirrors the standard model 
for non-CRN use cases, $y = \tb(x)+\epsilon$, 
where it is commonly assumed that all $\epsilon$ are 
independent Gaussian \textit{variable} realizations.
With CRN, the ``noise" terms $\epsilon_s(x)$ are independent 
Gaussian \textit{process} realizations. 
Secondly, $k_\epsilon(x,x')$
dictates the covariance in differences from the target at $x$ 
and $x'$ induced by CRN, we discuss
our choice below. 
Thirdly, $k_\epsilon(x,x')$ is typically a parametric function whose hyperparameters 
are learnt from multiple realizations, $\epsilon_1(x), \epsilon_2(x),...$, of a single GP and 
each seed may be viewed as a task in a multi-task model.
This differs slightly from other multi-task models commonly used for multi-fidelity
optimization \citep{swersky2013multi,poloczek2017multi}, or for multi-objective
optimization \citep{picheny2015multiobjective}, 
where one task is not necessarily the same as others and a unique GP 
model for each task may be more suitable.
However, because all $\epsilon_s(x)$ come from a single common GP, the kernel $k_\epsilon(x,x')$ 
must have the flexibility to model how the objective for any seed may differ from the target. 
We assume a decomposition of the difference functions, $\epsilon_s(x)$, 
into three parts: a constant offset~$o_s$, a bias function~$b_s(\cdot)$, and white noise~$w_s(\cdot)$:
\begin{eqnarray}
\theta(x,s)\,\,\, &=& \,\,\,\tb(x) + \epsilon_s(x)\, \,\,= \,\,\,\tb(x) + o_s + b_s(x) + w_s(x).
\end{eqnarray}{}
Firstly, to capture the 
notion that some seeds may result in scenarios that are ``easy" and others ``hard" for all inputs $x$, 
$\epsilon_s(x)$ may contain a global offset modeled by the constant kernel,
\begin{equation}
o_s(x) \sim \GP(0, k(x,x')=\eta^2),
\end{equation}
where the sample function is constant for all $x$ and hence denoted by $o_s\sim N(0, \eta^2)$. Secondly, to capture 
the notion that similar solutions should have similar outputs given the same seed, we include a ``bias'' function
modelled with another Mat\'ern or squared exponential kernel,
\begin{equation}
b_s(x) \sim \GP\left(0, k(x,x')=k_b(x,x')\right).
\end{equation}
Thirdly, to capture any other effects not modelled by $o_s$ and 
$b_s(x)$, such as discontinuities, we follow 
\cite{chen2012effects} and \cite{xie2016bayesian} and 
include a realization of white noise
\begin{equation}
w_s(x) \sim \GP\left(0, k(x,x')=\delta_{x'x}\sigma_w^2\right).
\end{equation}
Therefore, this functional form of $\theta(x,s)$ is a realization of the Gaussian process 
\begin{eqnarray}
\theta(x,s) &\sim& \GP\big(\bar\mu,\,\, \ktb(x,x') + \delta_{ss'}(\eta^2 + k_b(x,x') + \sigma_w^2\delta_{xx'} )\big) \\
&=& \GP\big(\bar\mu,\,\, k(x,s,x',s')\big).
\end{eqnarray}
See Figure~\ref{fig:gen_model} for example realizations. Although this is a general model, to simplify parameter learning 
in practice we assume parameter sharing between $\ktb (x,x')$ 
and $k_b(x,x')$ such that a CRN model has only two more hyperparameters
than its corresponding non-CRN model. We discuss in more detail in Section
\ref{sec:hypers}. For the rest of this section, we assume that all
kernels are known functions and the unknown $\theta(x,s)$ are to be inferred.
\ifwithpics
\begin{figure}[t]
    \centering
    \includegraphics[width=0.8\textwidth]{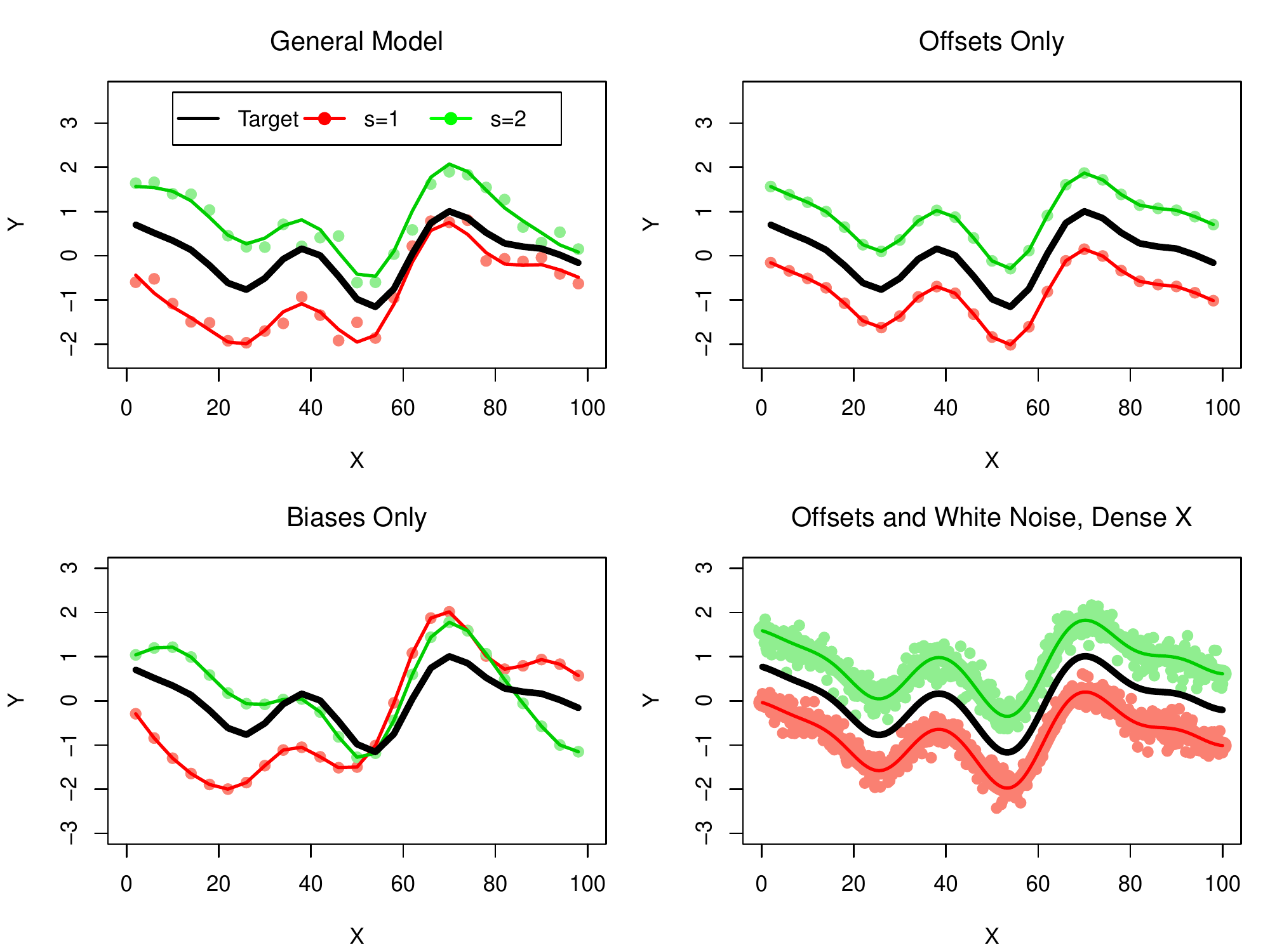}
    \caption{Samples from the generative model. In all plots, lines show
    $\tb(x)$ and $\tb(x)+o_s+b_s(x)$ (no white noise), points show $\theta(x,s)$
    (including white noise). Left plots: an algorithm must evaluate multiple seeds to find optimum.
    Right plots: an algorithm can optimize one seed to find $\argmax \tb(x)$.}
    \label{fig:gen_model}
\end{figure}
\fi
%
%
%
%
%
%
%
%
%
%
%
\subsection{Inferring the Objective $\theta(x,s)$} \label{sec:inf_simout}
We denote an observation at time $n$ as $(x^n, s^n, y^n)$, the sequence of observed
solutions as $(x^1,...,x^n)=X^n$, the sequence of observed seed values as $S^n$ and the sequence of
input pairs, $\tilde x^i = (x^i, s^i)$, as $(\tilde x^1,,...,\tilde x^n)=\tXn$. The vector of observed outputs is denoted $(y^1,...,y^n)=Y^n$. And, abusing notation, we also treat these as sets, e.g., $\tilde x \in \tXn$, and use both $(x,s)$ and $\tilde x$ interchangeably to represent an input pair.
The dataset of observed inputs and outputs we denote $D^n=((\tilde x^1,y^1), ..., (\tilde x^n,y^n))$.
Inferring the underlying realization of $\theta(x, s)$ 
can be done analytically using the  Bayesian update equations
for multivariate Gaussian random variables,
\begin{eqnarray}
\theta(x,s)|D^n &\sim& \GP\big(\,\mu^n(x,s), \,\, k^n(x,s,x's')\,\big)\nonumber\\
\mu^n(x,s) &=&  \mu^0(x,s) - k^0(x,s, \tilde X^n) K^{-1} (Y^n- \mu^0(\tilde X^n)) \label{eq:GPn}\\ 
k^n(x,s, x',s') &=&  k^0(x,s,x',s') - k^0(x,s, \tilde X^n) K^{-1} k^0(\tilde X^n, x',s') \label{eq:GPk}
\end{eqnarray}
where $k^0(x,s,x's')$ is any positive semi-definite kernel over $X\times \Np$. The matrix
$K=k^0(\tXn, \tXn)$ is the generative covariance for $Y^n$. For the rest
of this work, we use the shorthand $\E[1]$
. Note that there
is no added identity matrix as in Equation~(\ref{eq:genYiid}),
thus the model assumes deterministic outputs for any given input pair $(x,s)$.
%
%
%
At first, this may appear at odds with the white-noise assumption.
The posterior mean predicts a sum of GP realizations $\mu^n(x,s)=\E_n\left[\tb(x)+o_s+b_s(x)+w_s(x)\right]$. 
White noise has zero spatial correlation;
at observed input pairs, $(x^i,s^i)\in\tXn$,  the predicted
white noise realization is informed by data and $\E_n[w_{s^i}(x^i)]\neq 0$ (almost surely),
while at unobserved input pairs, it is not informed by data and $\E_n[w_{s}(x)]=0$. As a result, the posterior mean discontinuously interpolates the data as shown in Figure~\ref{fig:Example_GP_KG}.



\subsection{Inferring the Target $\tb(x)$}\label{sec:inf_obj}
The model of $\theta(x,s)$ 
and collected data is over the acquisition space $X\times\Np$
while the aim of the optimization is to maximize $\tb(x)$ over solution space 
$X$. The target 
is the objective averaged over infinite seeds and therefore the GP model of $\theta(x,s)$
averaged over infinite seeds induces another GP for the target $\tb(x)$ as follows.
\begin{lemma}\label{prop:thetabar_gp}
For any given kernel over $X\times \Np$ that is of
the form $k_{\tb}(x,x') + \delta_{ss'}k_\epsilon(x,x')$, 
and a dataset of $n$ input-output triplets $D^n$, the posterior over the  target is a Gaussian process given by
\begin{eqnarray}
\tb(x)|D^n  &\sim & \GP(\mu^n_{\tb}(x), k^n_{\tb}(x,x')) \\
\mu^n_{\tb}(x) &=&  \mu^n(x,s') \label{eq:inf_obj}\\
k^n_{\tb}(x, x') &=&  k^n(x,s', x', s'')
\end{eqnarray}
where $s', \,s''\in \Np\setminus S^n$ with $s'\neq s''$  are any two unobserved unequal seeds.
\end{lemma}
The intermediate steps and proof are given in the Electronic Companion EC.0.1. For the sake of a simple notation,
we assume that seeds are labeled by positive integers, and let $s'=0$ and
$s''=-1$. Then  $\mu^n(x,0)$ is the posterior expectation of the target $\tb(x)$. 
%
%
%
%
%
%


%
%



\section{Knowledge Gradient for Common Random Numbers} \label{sec:acqfun}

\subsection{Acquisition Function}
Evaluations of $\theta:X\times \Np\to \R$ are collected in order to optimize $\tb:X \to \R$. 
Given a joint model of both functions, the acquisition function quantifies the benefit of a new
hypothetical observation at $(x,s)\in \tilde X$. This function is then optimized to
obtain the best $(x,s)\n1$ and the objective is evaluated $y\n1=\theta(x\n1, s\n1)$. 
The surrogate model is defined over the space of non-negative seeds $X\times \{0,1,...\}$, 
the model of the target is over $X \times \{0\}$ while the objective, and acquisition,
is over $X\times \{1,2,..\}$. 
Therefore we require a `correlation aware' acquisition function that computes the benefit
of a sample at $(x,s)\n1$ for $s\n1>0$ by measuring changes
in the model at other locations $(x',0)\neq (x,s)\n1$. 
This requirement excludes certain acquisition functions in their unmodified form
such as Expected Improvement \citep{jones1998efficient}, Upper Confidence Bound
\citep{srinivas2009gaussian} and Thompson sampling \citep{kandasamy2018parallelised}.
Two popular families of acquisition functions that 
naturally account for how the whole surrogate model changes include Entropy Search 
\citep{ES1_villemonteix2009informational}, and Knowledge Gradient \citep{frazier2009knowledge}. 
Knowledge Gradient quantifies the benefit of a new hypothetical point $(x,s,y)\n1$ as the expected
incremental increase in the predicted outcome 
for the user, peak posterior mean 
$\E[\max_x \mu\n1(x)-\max_x\mu^n(x)|D^n, x\n1]$. 
In this work we adopt the Knowledge Gradient for its principled
value of information-based approach and provable performance guarantees.

In our setting the value of information is the 
expected increase in the predicted peak of the target, $\max \mu\n1(x,0) - \max \mu^n(x,0)$, 
caused by a new sample $y\n1$ at $(x,s)\n1$. 
The Knowledge Gradient for Common Random Numbers, $\CRNKG_n:\tilde X \to \R^+$, is given by
\begin{eqnarray} \label{eq:KGCRN}
\CRNKG_n(x,s) &=& \E_n\bigg[\max_{x'\in X}\mu\n1(x',0) - \max_{x''\in X} \mu^n(x'',0)\bigg| (x,s)\n1=(x,s)\bigg] \\
&=& \E_n\bigg[\max_{x'\in X}\mu^n(x',0) + \tilde\sigma^n(x',0;x,s)Z- \max_{x''\in X} \mu^n(x'',0)\bigg]
\end{eqnarray}
where, conditioned on $D^n$, the expectation is only over $Z\sim N(0,1)$ and 
$$\tilde\sigma^n(x,0;(x,s)\n1) = \frac{k^n(x,0,(x, s)\n1) }{ \sqrt{k^n((x,s)\n1,(x,s)\n1)}  } .$$
A full derivation can be found in multiple previous works \citep{frazier2009knowledge, pearce2017bayesian}.
The next input to the objective, $(x,s)\n1$, is determined by optimizing the
above acquisition function
$(x,s)\n1 = \argmax_{x,s}\CRNKG_n(x,s).$
Evaluation of $\CRNKG$ is the expectation of a maximization and can be evaluated analytically
when $X$ is a finite set. For the general case, approximations are required that we discuss in Section \ref{sec:evalKG}. 
Moreover, we show in Section~\ref{sec:optimKG} how to cheaply compute  $\argmax_{x,s}\CRNKG_n(x,s)$.

The acquisition space, $X\times \Np$, contains an infinite number of seeds. However as a 
result of the assumed form of the GP, the posterior mean and correlation
are identical for all \emph{unobserved} new seeds $s\in \Np\setminus S^n$. Thus,
the value under the acquisition criterion is
identical for all new seeds, $\CRNKG_n(x,s)=\CRNKG_n(x,s')$ for all $s,s'\in \Np \setminus S^n$.
Hence, it suffices to consider the acquisition criterion on all observed seeds $s\in S^n$ and
only a single new seed $s=\max\{ S^n\}+1$. 
Over multiple iterations, new seeds may be evaluated and added to the set of observed seeds and the acquisition space grows accordingly by always including one new seed. 
Note that the acquisition criterion is maximized jointly over the old and new seeds. In particular, no heuristics or user input is used to make the exploration-exploitation trade-off over old and new seeds.

A connection can be drawn between our algorithm and recent work on multi-information source optimization \citep{swersky2013multi,poloczek2017multi}. At a given iteration, each seed in the acquisition space may be viewed as an information source and $s=0$ is the target, and 
a user must choose a solution $x$ and an information source
$s$ in order to optimize a target $s=0$. 
However in the CRN case, the target itself cannot be observed, all sources have equal budget consumption, and the number of available sources is infinite.

\ifwithpics
\begin{figure}[t]
    \centering
    \includegraphics[width=0.9\textwidth]{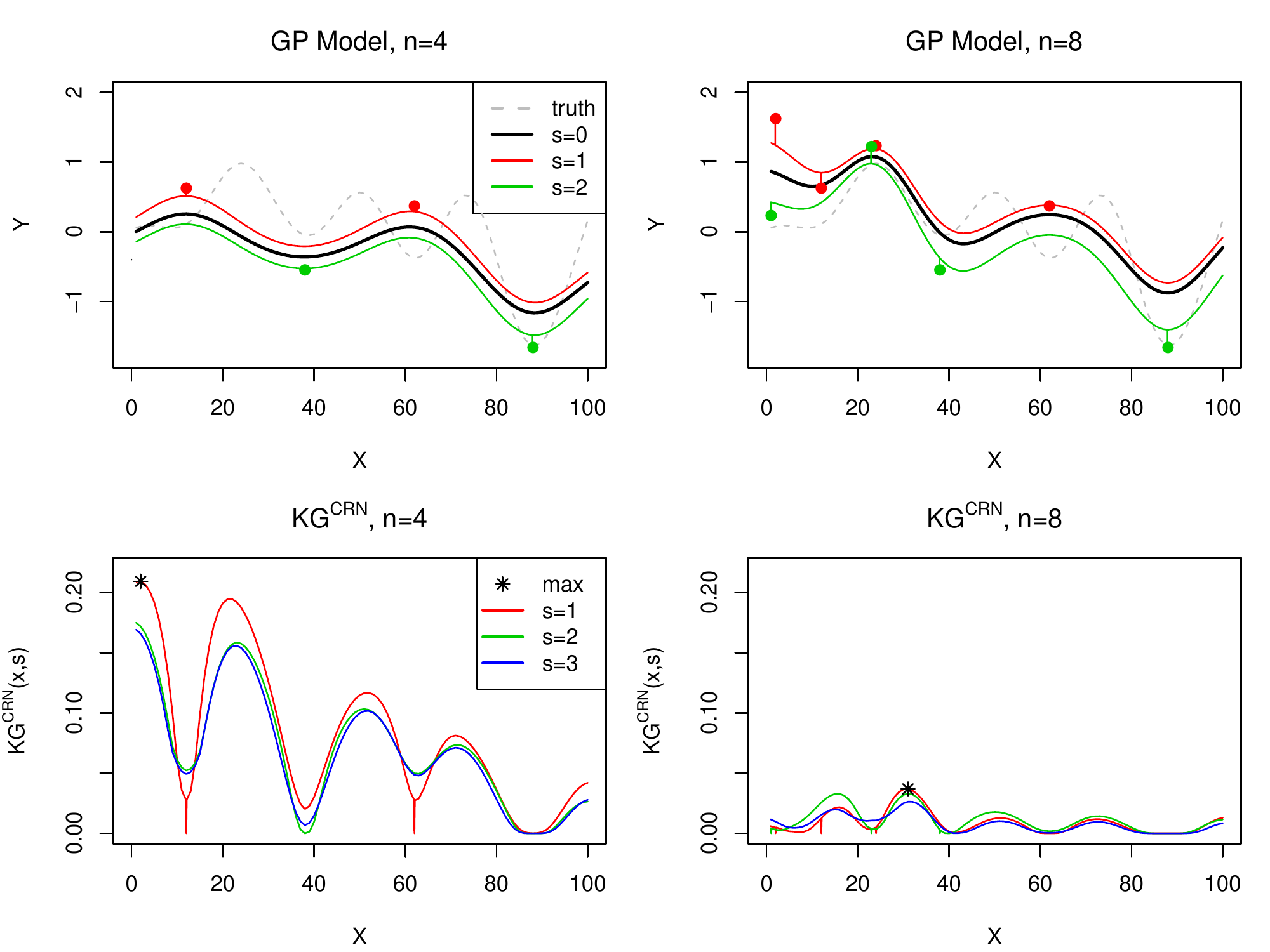}
    \caption{(top) The GP model with offsets, bias functions and 
    white noise. 
    (bottom) $\CRNKG$ after 4 initial points on
    seeds $s=1,2$ (left) and an added 4 sequential points by $\CRNKG$ (right).
    All new points were allocated to seeds $s=1,2$ and the next
    point will be allocated to $s=1$.}
    \label{fig:Example_GP_KG}
\end{figure}
\fi

\subsection{Implementation Details} \label{sec:implementation}

In Section \ref{sec:method} we assume that $k_{\tb}(x,x')$ and
$k_{\epsilon}(x,x')$ are known while in practice they require
hyperparameters estimated from data.  
Also in Section \ref{sec:acqfun} we assume $\CRNKG_n(x,s)$
can be evaluated and maximized. These practical issues apply to non-CRN and CRN algorithms,
however the CRN model has both more hyperparameters and a larger acquisition space.
Ideally, incorporating CRN should not require significantly more computational resources
and we discuss such solutions below.

\subsubsection{Gaussian Process Hyperparameters.} \label{sec:hypers}
In this work we assume that the target is modeled with the 
popular squared exponential (SE) kernel
$$ \ktb(x,x') = \sigma_{\tb}^2\exp(-(x-x')\trans L (x-x')/2)$$ 
where $L=\text{diag}(1/l_1^2,...,1/l_d^2)$ is a
diagonal matrix of inverse length scales. We also assume that
the bias functions come from a squared exponential kernel 
$k_b(x,x') = \sigma_{b}^2\exp(-(x-x')\trans L (x-x')/2)$ that
shares the diagonal matrix $L$. 
The constant kernel and white noise kernel each have a single parameter
$\eta^2$ and $\sigma^2_w$. 
The constant kernel, over $X$, models infinitely long range correlation in differences while
the white noise kernel models infinitely short range. Therefore the bias kernel only needs to model
intermediate ranges. When determining an intermediate range, one option is to learn hyperparameters
for the bias kernel, however in preliminary testing this led to unstable model fitting.
Instead we simply share such bias kernel hyperparameters, i.e. length scales, with the kernel of the
target. This greatly simplifies model learning and still allows the GP to capture
the necessary intermediate range correlation. For any $k_{\tb}(x,x')$, 
one may use $k_b(x,x')\propto  k_{\tb}(x,x')$ where the ratio is a hyperparameter. 
Therefore, the only design choice to be made for the CRN model is $k_{\tb}(x,x')$.
In total, the model has parameters 
$L, \sigma_{\tb}, \eta^2, \sigma_b^2, \sigma_w^2$, two more than
a non-CRN model. All parameters are learnt by first maximizing
the marginal likelihood for a non-CRN model (i.e. clamping $\eta^2=\sigma_b^2=0$), using multi-start gradient ascent.
This is followed by fine tuning  the hyperparameters of the difference kernel, $\eta^2$, $\sigma_b^2$, $\sigma_w^2$,
with the constraint 
$\eta^2+\sigma_b^2+\sigma_w^2 = \sigma^2_{w, \text{non-CRN}}$
such that the variance of the difference functions is the same as the
variance of the noise in the independent model. This is a single Nelder-Mead local hill-climb over a two dimensional optimization.
In a final step, we fine-tune all hyperparameters simultaneously (one more gradient ascent).
Overall, the only difference between fitting a non-CRN model and a CRN model is
in the added  two local optimization steps.
For details, see the Electronic Companion~\ref{EC:hp_optim}. 
In future work, especially with more complex models, we will study a Bayesian treatment
of the hyperparameters: such an approach can improve algorithm performance especially
for very small budgets when hyperparameters are most uncertain.

\subsubsection{Evaluation of $\CRNKG_n(x,s)$.} \label{sec:evalKG}
The acquisition function, Equation~(\ref{eq:KGCRN}),
is a one-step look-ahead expected peak posterior mean, an expectation of maximizations over $X$. 
This may be evaluated analytically when $X$ is a feasibly small finite
set using Algorithm 1 from \cite{frazier2009knowledge}.
Alternatively, when $X$ is a continuous set, one may replace the expectation over the
infinite $Z$ with a Monte-Carlo average.
For each $Z$ sample, the inner maximization is performed over $X$ numerically, 
yielding a stochastic unbiased estimate of $\CRNKG_n(x,s)$ \citep{wu2017bayesian}.

In this work, we follow \cite{poloczek2017multi} 
and \cite{xie2016bayesian} that use a deterministic approximation. This allows us
to reliably test a conjecture and allows direct comparison with prior work
both described in Section~\ref{sec:comparePWKG}.
The inner maximization over $X$ may be replaced with a smaller random
subset $A$ that is frozen between iterations thus approximating $\CRNKG$ with
\begin{eqnarray} \label{eq:KG_discrete}
\CRNKG_n(x,s;A) = \E_n\bigg[\max_{x'\in A\cup \{x\}}\mu^n(x',0) + \tilde\sigma^n(x',0;x,s)Z 
- \max_{x''\in A\cup \{x\}} \mu(x'',0) \bigg].
\end{eqnarray}
We desire a discretization, $A\subset X$, that is both dense around promising
regions in $X$ while still accounting for unexplored regions.
Thus, we propose to construct $A$ from a union of a latin hypercube over $X$ with $n$ points,
$A^n_{LHC}$, and random perturbations of previously sampled points 
$A^n_{P}=\{x^i+\gamma | x^i \in X^n\}$ where $\gamma \sim N(\underline{0},I)$
is Gaussian noise scaled for the application at hand.
Finally, we let $A^n=A^n_{LHC}\cup A^n_{P}$.

\subsubsection{Optimization over the Acquisition Space.} \label{sec:optimKG}
Typically, acquisition functions are multi-modal  functions
over $X$ and maximized by multi-start gradient ascent.
For  $\CRNKG_n(x,s)$, the acquisition space is larger
$\tilde X_{acq}^n = X\times \{1,...,\max S^n+1\}$, suggesting $\CRNKG_n(x,s)$
needs to be optimized over $X$ for each $s$. 
However, recall the fundamental CRN modelling assumption that all seeds
have the same latent $\tb(x)$. As a result, $\CRNKG_n(x,s)$
for each seed often has peaks and troughs in similar
locations, see Figure~\ref{fig:Example_GP_KG}.
Therefore, to maximize $\CRNKG_n(x,s)$, one may use the same multi-start gradient ascent
method for a non-CRN method where instead
each start is allocated to a random seed $s_i$ and optimizes $x$ 
over $X\times \{s_i\}$. Using the best point so far, $(x_{ga},s_{ga})$,
the same $x_{ga}$ is evaluated for all seeds to find $s_{final}$
and one run of gradient ascent over $X\times \{s_{final}\}$ starting from $x_{ga}$ yields $x_{final}$. 
Thus, the only difference in computational cost of acquisition optimisation 
between a non-CRN method optimizing over $X$ and a CRN method optimizing over $X\times \Np$
is in the final phase from $(x_{ga}, s_{ga})$ to $(x_{final}, s_{final})$.

\begin{algorithm}
\caption{The $\CRNKG$ Algorithm.\label{alg:CRNKG}}
\begin{algorithmic}[1]
\REQUIRE $\theta(x,s)$, \,$X$,\, $n_{init}$, \,$N$,\, $k_{\tb}(x,x')$, \,
method to evaluate $\E[\{\max_{x'}a(x')+b(x',x)Z\}]$ and $\nabla_x\E[\{\max_{x'}a(x')+b(x',x)Z\}]$, \,
\texttt{Optimizer}() over $X\times \Np$

\STATE $\tilde X^{n_{init}}\leftarrow$ $n_{init}$ sampled points by LHC over $X\times \{1,2,3,4,5\}$ 
\STATE $Y^{n_{init}}\leftarrow \theta(\tilde X^{n_{init}})$

\FOR{$n=n_{init}$ \TO $N-1$ } 

\STATE  $\mu^n(x,s), k^n(x,s,x',s')\leftarrow \GP\big(\theta(x,s) \big|\tilde X^n, Y^n, L, \sigma_{\tb}^2, \eta^2, \sigma_b^2, \sigma_w^2\big)$ with MLE hyperparameters
\STATE  $\CRNKG_n(x,s)\leftarrow \E[\{\max_{x'}\mu^n(x',0)+\st^n(x',0,x,s)Z\}] -\max_{x''}\mu^n(x'',0) $ and gradient w.r.t. $x$
\STATE  $(x,s)\n1\leftarrow$\texttt{Optimizer}($\CRNKG_n(x,s)$)
\STATE  $y\n1\leftarrow \theta(x\n1,s\n1)$ 
\STATE  $\tilde X\n1, Y\n1\leftarrow  (\tilde X^n, (x,s)\n1), (Y^n, y\n1)$

\ENDFOR

\STATE  $\mu^N(x,s) \leftarrow \GP\big(\theta(x,s) \big|\tilde X^N, Y^N, L, \sigma_{\tb}^2, \eta^2, \sigma_b^2, \sigma_w^2\big)$ with MLE hyperparameters

\RETURN $x_r^N = \amax{x}\mu^N(x,0)$
\end{algorithmic}
\end{algorithm}


\subsection{Algorithm Properties}

The acquisition benefit obtained by sampling solution $x$ with seed $s$ is the expected gain in
the quality of the best solution that can be selected given all the available information. In this regard, the $\CRNKG$
is one-step Bayes optimal by construction. 
The following observation is trivial yet worth highlighting: standard Knowledge
Gradient (KG) is reproduced by constraining $\CRNKG$ to only acquire data for a new seed in each iteration. Thus, we have
\begin{equation} \label{eq:CRNgreaterKG}
\max_{x,s\in\Np}\CRNKG_n(x,s) \geq \max_{x,s\in \Np\setminus S^n}\CRNKG_n(x,s)=\max_x \text{KG}(x)
\end{equation}
and sampling without CRN is a lower bound on the acquisition benefit achievable by $\CRNKG$.

Given an infinite budget, it is a desirable property for any algorithm 
to be able to discover the true optimum $x^{OPT} = \amax{x\in X}\tb(x)$ (assuming there is only one optimizer). 
Here we give an additive bound on the loss when applying $\CRNKG$
to a finite subset, $A$, of continuous space $X$. Let $k_{\tb}(x,x')$ be a
Mat\'ern class kernel, and $d=\max_{x'\in X}\min_{x\in A}\text{dist}(x,x')$ 
the largest distance from any point in the continuous domain $X$ 
to its nearest neighbor in $A$.
\begin{theorem} \label{thm:KG_assymptotic}
Let $x^N_r\in A$ be the point that $\CRNKG$ recommends in iteration $N$. For each $p\in [0,1)$, there is a constant
$K_p$ such that with probability $p$
$$\lim_{N\to\infty}\tb(x_N^{r})> \tb(x^{OPT})-K_pd$$
holds.
\end{theorem}
The proof is given in the Electronic Companion EC.0.2. 
Note that this establishes consistency for the finite case as $A=X$ and $d=0$.
Clearly, this bound is conservative as $A$ is randomized at each iteration to avoid ``overfitting'' and $\CRNKG$ recommends
the best predicted solution in $X$, not restricted to~$A$.


\section{Comparison with Previous Work} \label{sec:prev_methods}
We first show how to recover the generative model considered by \cite{xie2016bayesian} and 
\cite{chen2012effects} as a special case of our proposed model. We then discuss the method of 
\cite{xie2016bayesian} that also extended Knowledge Gradient to account for common random numbers.
\subsection{Compound Sphericity}
If there are no bias functions, $k_b(x,x')=0$, the differences
kernel reduces to 
$k_\epsilon(x,x')=\eta^2 + \sigma_w^2\delta_{xx'}$ and
each difference function $\epsilon_s(x)$ is an offset and 
white noise. Thus, the differences matrix
$k_\epsilon(X^n, X^n)$ is $\eta^2+\sigma_w^2$ 
on the diagonal and constant $\eta^2$ for all off-diagonal terms, this matrix composition is referred to as compound sphericity.
The correlation in differences may be written as 
$\rho=\eta^2/(\eta^2+\sigma_w^2)$. 
Let $\Delta^n = Y^n-\mu^0(\tXn)$ and 
$\One_s=\One_{s\in S^n}\in \{0,1\}^n$ be a binary masking vector. $\One_x$ is defined analogously.
Then the posterior mean 
has the following simple form:
\begin{eqnarray}
\mu^n(x,s) &=&  \mu^0(x) - (\ktb(x,X^n) + \eta^2\One_s + \sigma_w^2\One_s \One_x )K^{-1} \Delta^n  \nonumber \\ 
&=& \underbrace{\ktb(x,X^n) K^{-1} \Delta^n}_{\mu^n(x,0)}
\,\, +\,\, \underbrace{\eta^2\One_s K^{-1} \Delta^n}_{\text{independent of $x$}}
\,\, +\,\, \underbrace{\sigma_w^2\One_s \One_x K^{-1} \Delta^n}_{=0\text{ except for }(x^i,s^i)\in \tXn} \nonumber   \\
&=& \mu^n(x,0)\,\,+\,\, A_s \,\, +\,\, B_s\One_{(x,s)\in\tXn}  \label{eq:mu_cs}
\end{eqnarray}
%
%
and the posterior mean function for a given seed, $s>0$, differs from the 
target, $s=0$, by two additive terms. The first is a constant $A_s$ and the second
is non-zero for singletons ${(x,s)\in\tilde X^n}$. This leads 
to the following two Lemmas, both cases correspond to the second additive
term equating to zero.
Firstly, if there is no white noise $(\sigma_w^2=0)$ then for all seeds
$\epsilon_s(x)=o_s$ is only a constant offset and a user may simply
optimize a single seed to learn $\argmax \tb(x)$.
This corresponds to compound sphericity with full correlation, $\rho=1$, and
may be viewed as a ``best case'' scenario for CRN.

\begin{lemma} \label{lem:cs_cor}
Let the function $\theta(x,s)$ be a realization of a Gaussian 
process with compound sphericity with full correlation, $\rho=1$. 
Then for all $s\in \Np$, the posterior mean functions have the same optimizer as the target estimate
$$\argmax_{x\in X} \E_n[\tb(x)] = \argmax_{x\in X} \mu^n(x,s') \quad\quad\forall s'\in \Np.$$
\end{lemma}
\proof{Proof}
By setting $\sigma_w^2=B_s=0$ in Equation (\ref{eq:mu_cs}), the posterior
means for all seeds differ by only an additive constant, $A_s$, therefore 
the maximizer of any two seeds is the same and by Lemma~\ref{prop:thetabar_gp} the same maximizer as the estimate of $\E_n[\tb(x)]$.
\Halmos
\endproof
%
%
%

Secondly, when there is white noise and the set of solutions $X$ is large and dense,
a user may simply optimize a single seed to learn $\argmax \tb(x)$ as above.
\begin{lemma} \label{lem:cs_infX}
Let the function $\theta(x,s)$ be a realization of a Gaussian process with
compound sphericity over a continuous set of solutions $X$, 
then for all $s\in \Np$, the posterior mean functions have the same optimizer excluding
past observation singletons $\tXn$
$$\argmax_{x\in X\setminus X^n} \E_n[\tb(x)] = \argmax_{x\in X\setminus X^n} \mu^n(x,s') \quad\quad\forall s'\in \Np.$$
\end{lemma}
\proof{Proof}
By excluding singletons $x\in X^n$, the second additive term in Equation (\ref{eq:mu_cs}) vanishes
$(B_s\One_{(x,s)\in\tXn}=0)$. The posterior
means for all seeds differ by only an additive constant, $A_s$, therefore 
the maximizer of any two seeds is the same and by Lemma 
\ref{prop:thetabar_gp} the same as $\E_n[\tb(x)]$.
\Halmos
\endproof
%
%
%
%

The right column of Figure~\ref{fig:gen_model} illustrates example functions for these cases
and top row of Figure~\ref{fig:Example_GP_KG} shows how the posterior mean is discontinuous at evaluated points.
If there are no bias functions and these discontinuities are excluded,
the posterior mean has the same shape for all seeds.
Consequently, for a function that is a realization of a GP with the compound spheric
noise model, if there is high correlation or a large and dense number of
solutions $X$, allocating samples to a single seed can be much
more efficient than allocating to multiple seeds.
This result agrees with those found by \cite{chen2012effects}:  in the case $\rho=1$
with data collected on seed $s=1$, the intercept of the function $\tb(x)$ is less accurately
known while derivatives $\nabla_x\tb(x)$ are more accurately known. This is because in 
the $\rho=1$ case, the generative modelling assumption imposes the functional form as 
$\theta(x,s)=\tb(x)+o_s$ implying $\nabla_x\theta(x,s)=\nabla_x\tb(x)$.
%
It is due to the presence of the \textit{bias} functions, $b_s(x)$, 
that the optimizer of one seed, $\argmax_{x}\theta(x,s)$, is not an accurate
estimate of the optimizer of the target function, $\argmax_{x}\tb(x)$, and
an optimization algorithm must evaluate multiple seeds.

Next, in Lemma~\ref{thm:none_left}
we show that if all solutions of a finite set $X$ have been evaluated there
is no more acquisition benefit according to $\CRNKG$, the optimizer is known 
even though its underlying value is unknown. 
\begin{lemma}\label{thm:none_left}
Let $\theta(x,s)$ be a realization of a Gaussian process with the compound
spheric kernel and $\rho=1$. Let $X=\{x_1,...,x_d\}$ and evaluated points
$\tilde X^n=\{(x_1, 1),...,(x_d, 1)\}$,
then for all $(x,s)\in X\times \Np$, there is no more value of any measurement
\begin{equation}
\CRNKG_n(x,s)=0
\end{equation}
and the maximizer $\argmax{x}\tb(x)$ is known.
\end{lemma}
Proof is given in the Electronic Companion EC.0.3.
%
%
%
%

Next, $\CRNKG$ may be evaluated according to the method proposed by
\cite{KGCP_scott2011correlated}. The method discretizes the inner
maximization over $X$ with past evaluated points, $X^n$, and the new proposed point
so that the integral over $Z$ is analytically tractable. This
may be viewed as a noise-generalized Expected
Improvement (EI) because it reduces to  EI \citep{jones1998efficient} when outputs are deterministic. By augmenting
this KG evaluation method with the ability to choose the seed, in the full
correlation case it is guaranteed to never evaluate a new seed and
the $\CRNKG_n(x,s)$ function also simplifies to EI applied to seed $s=1$.
\begin{lemma}\label{prop:KGCS_never_new}
Let $\theta(x,s)$ be a realization of a Gaussian process with the compound spheric kernel with $\rho=1$.
Let $X\subset \R^d$ be the set of possible solutions, $\tXn=\{(x^1,1),...,(x^n,1)\}$ be the set of sampled
locations and $X^n=(x^1,...,x^n)$. Define
\begin{eqnarray}
\CRNKG_n(x,s; A) = \E_n\bigg[\max_{x'\in A\cup \{x\}}\mu\n1(x',0)- \max_{x'\in A\cup \{x\}} \mu^n(x',0)\bigg|(x,s)\n1=(x,s)\bigg].
\end{eqnarray}
Then for all $x \in X$
$$\CRNKG_n(x,1; X^n)>\CRNKG_n(x,2; X^n)$$
and therefore $\max_x\CRNKG_n(x,1; X^n)>\max_x\CRNKG_n(x,2; X^n)$ and seed $s=2$ will never be evaluated.
Further 
$$\CRNKG_n(x,1; X^n) = \E\big[\max\{ 0, y\n1 - \max Y^n\}\big |D^n, x\n1=x, s\n1=1\big].$$
\end{lemma}
The proof is given in the Electronic Companion EC.0.3. 

In the more general case, evaluating $\CRNKG_n(x,s)$ by any method,
when using compound spheric with either full correlation
or in a continuous domain $X$, we conjecture that the
true myopically optimal behaviour is
to never go to a new seed,
$$ \max_{x\in X, s_{old}\in S^n}\CRNKG_n(x,s_{old}) >  \max_{x\in X, s_{new}\notin S^n}\CRNKG_n(x,s_{new})$$
and a new seed $s\notin S^n$ will never be sampled.
However, the above inequality cannot be proven because $\max_{x\in X}\CRNKG(x,s)$ has no analytic
expression and must be found numerically via gradient ascent algorithms.
(Note that $\CRNKG_n(x,s_{old}) >  \CRNKG_n(x,s_{new})$ is not true in general,
$x^i\in X^n$ are counter examples.) Therefore we numerically demonstrate
this conjecture in Section \ref{sec:numerical}.

However, this conjectured behaviour
comes with the risk that if the modelling assumption is incorrect
for a given application, the algorithm will try to optimize
a single seed and never find the true optimum of $\tb(x)$.
We observe this phenomenon in Section \ref{sec:numerical} 
where compound sphericity on a continuous search space encourages greedy resampling 
of only observed seeds. However this does not happen with the inclusion of bias functions,
bias functions allow for more intelligent modelling of noise structure that can then
be exploited more appropriately.


\subsection{Comparison with Knowledge Gradient with Pairwise Sampling\label{sec:comparePWKG}}
The method proposed by \cite{xie2016bayesian} was also an extension of Knowledge Gradient
to use common random Numbers. For the generative model, the
method assumes that $\bar\theta(x)$ is a realization of a GP and
considers compound spheric covariance for difference functions.
For acquisition, the standard Knowledge Gradient
acquisition function quantifies the value of a single observation without CRN (on a new seed) and
this is extended with a second acquisition function that quantifies the value of a
pair of observations with CRN (on the same new seed). 
The acquisition space is thus $\tilde{X}^{PW} = \{X, X\times X\}$. The method switches between the
serial mode and the batch mode depending on which mode promises the larger value per
sample. Since the value of a pair cannot be computed analytically,
a lower bound is given by considering the \textit{difference}
between the pair of outcomes
\begin{eqnarray}\label{eq:KGPW}
\PWKG_n(x_i,x_j) &=& \frac{1}{2} \left( \E_n\bigg[\max_{x'\in X}\mu^n(x',0) + \tilde{\tilde\sigma}^n(x',0;x_i,x_j)Z- \max_{x''\in X} \mu^n(x'',0) \bigg]\right)\\
\tilde{\tilde\sigma}^n(x,0;x_i,x_j) &=& \frac{k^n(x,0,x_i,s\n1) - k^n(x,0,x_j,s\n1)
}{
\sqrt{k^n(x_i,s\n1, x_i,s\n1) + k^n(x_j,s\n1, x_j,s\n1) - 2k^n(x_i,s\n1, x_j,s\n1)}}
\end{eqnarray}
where $s\n1=n+1$ is a new seed and $\PWKG_n(x,x')$ is optimized over $X\times X$. 
Note we have adapted the notation from the original work
where the seed is not an explicit argument to the formulation presented in this work.
In the original work, numerical evaluation of $\PWKG$ is performed by discretizing
the inner maximization, as discussed in Section \ref{sec:evalKG}. One call to $\PWKG$
requires evaluating both $k^n(x,0,x_i,s\n1)$ and $k^n(x,0,x_j,s\n1)$ for each $x$ and
is thus more expensive than one call to KG or $\CRNKG$.

In the large $|X|$ setting, it is efficient to use GP regression, with compound sphericity 
in the high $\rho$ setting it is efficient to use CRN. Within both of these regimes,
it is doubly beneficial to revisit old seeds as implied by both Lemmas \ref{lem:cs_cor} and \ref{lem:cs_infX}. 
Therefore, the Knowledge Gradient with Pairwise Sampling combines an acquisition
procedure that can only sample new seeds with a differences model for which
it is efficient to only sample old seeds.
From a value of information perspective,
both serial and batch modes of $\PWKG$ yield equal
or lower value of information than sequential allocation by $\CRNKG$.

\begin{lemma} Let $D^n$ be a dataset of observation triplets. For a Gaussian process with a kernel of
the form $k_{\tb}(x,x')+\delta_{ss'}k_\epsilon(x,x')$, 
the expected increase in value after two steps allocated according to $\CRNKG$ is at least as big as two
steps allocated according to $\PWKG$,
\begin{eqnarray*}
&&\E_n\left[\max_{x'}\mu^{n+2}(x',0) - \max_{x''}\mu^{n}(x'',0)\big|(x,s)\n1,(x,s)^{n+2}\sim \CRNKG\right] \\ 
&\geq&\E_n\left[\max_{x'}\mu^{n+2}(x',0) - \max_{x''}\mu^{n}(x'',0)\big|(x,s)\n1,(x,s)^{n+2}\sim \PWKG\right] 
\end{eqnarray*}
\end{lemma}
\proof{Proof}
The suboptimality of one or two steps of the serial mode of $\PWKG$ is clear by noting it is constrained to a
new seed, a subset of the same acquisition space considered by $\CRNKG$ as mentioned
in Equation (\ref{eq:CRNgreaterKG}). We focus on the suboptimality of one step of the batch mode
\begin{eqnarray}
&&\E_n\left[ \max_{x'}\mu^{n+2}(x',0) - \max_{x''}\mu^{n}(x'',0)\big|(x,s)\n1,(x,s)^{n+2}\sim \CRNKG\right]\nonumber \\
&=& \max_{(x,s)\n1} \E_n\bigg[ \max_{(x,s)^{n+2}} \E_{n+1}\big[\max_{x'}\mu^{n+2}(x',0)\big|(x,s)^{n+2}\big]
-\max_{x''}\mu^n(x'',0) \bigg|(x,s)^{n+1}\bigg]\nonumber  \\
&\geq& \max_{x\n1}\E_n\bigg[ \max_{x^{n+2}}  \E_{n+1}\big[\max_{x'}\mu^{n+2}(x',0)\big|x^{n+2}\big]
-\max_{x''}\mu^n(x'',0)\bigg|x^{n+1}, s\n1=s^{n+2}=n+1\bigg]\label{eq:dbl_seq_PW}\\
&\geq& \max_{x\n1,x^{n+2}} \E_n\bigg[\max_{x'}\mu^{n+2}(x',0)-\max_{x''}\mu^n(x'',0)\bigg| x^{n+1},x^{n+2}, s\n1=s^{n+2}=n+1\bigg]  \label{eq:max_batch_PW}\\
&\geq& \E_n\bigg[\max_{x'}\mu^{n+2}(x',0)-\max_{x''}\mu^n(x'',0)\bigg|(x^{n+1},x^{n+2})=\argmax{} \PWKG_n(x,x'), s\n1,s^{n+2}=n+1\bigg]  \nonumber \\
&=&\E_n\left[\max_{x'}\mu^{n+2}(x',0) - \max_{x''}\mu^{n}(x'',0)\big|(x,s)\n1,(x,s)^{n+2}\sim \PWKG\right]\nonumber 
\end{eqnarray}
where the first inequality is 
due to constraining the acquisition space to a new seed,
the second is by Jensen's inequality and the convexity of the max operator implying sub-optimality
due to batch pre-allocation, and the third inequality is due to the approximation
with differences used in $\PWKG$ as pairs are not allocated to maximize the true batch value.
\Halmos
\endproof
%

%

Sequentially allocating two singles to the same new seed is guaranteed to have higher value than a corresponding
batch mode pre-allocating a pair to a single seed as shown by Equations (\ref{eq:dbl_seq_PW}) and (\ref{eq:max_batch_PW}).
However the serial and batch mode of $\PWKG$  compute the value over
different subsets of the full acquisition space
and therefore the batch mode can return higher value per sample. 
%

Instead, we make explicit the domain for the objective function as both a decision variable $x$ and a seed $s$
and build a surrogate model and acquisition procedure over the same space. This approach has many advantages. 
Firstly there is no need to consider batches/pairs, reducing the search space for the
acquisition from $X\times X$, reducing the cost per call to the acquisition function, and increasing the theoretical value of information. Secondly the structure in
the noise, difference functions, can be more aggressively exploited
allocating budget to either a few seeds or many new seeds as necessary. 
Thirdly, the GP model allows a user to replace KG with any multi-fidelity/multi-information source \citep{huang2006sequential,poloczek2017multi}
or `correlation aware' serial acquisition procedure and a corresponding parallel batch acquisition
function is not required. 

On the other hand, when enabling resampling of old seeds, assuming compound sphericity incentivises 
sampling of old seeds. 
The $\CRNKG$ algorithm includes bias functions
enabling accurate modelling and the appropriate trade-off between old and new seeds. The $\PWKG$ 
does not encounter such pitfalls as it does not sample old seeds.

\section{Numerical Experiments} \label{sec:numerical}
We perform three sets of experiments, first using synthetic GP sample
functions and known hyperparameters, allowing perfect comparison of just the acquisition procedures.
The next two problems are taken from the SimOpt library (\url{http://simopt.org}),
the Assemble-to-order problem (ATO) and the Ambulances in a Square problem (AIS). The code for all 
experiments will be made public upon publication.

\subsection{Compared Algorithms and Variants}
We aim to investigate the empirical effects of including bias functions and the ability 
of the acquisition procedure to revisit old seeds whilst holding all other experimental 
factors constant. Therefore we consider the following five algorithms.
%
%
\begin{itemize}
    %
    \item[] {\bf Knowledge Gradient (KG):} A GP model with independent homoskedastic noise
    is fitted, $\eta^2=\sigma_b^2=0$, $\sigma_w^2>0$. Acquisition is according to $\CRNKG$ artificially
    constrained to a new seed.
    \item[]{\bf KG with Pairwise Sampling ($\PWKG$):} Proposed by \cite{xie2016bayesian}. 
    A GP with the compound spheric differences kernel is fitted $\sigma_b^2=0$,
    $\eta^2, \sigma_w^2\geq 0$. For acquisition, the value of a single sample is given by $\CRNKG$ and pairs by
    $\PWKG$, both are constrained to a new seed.
    \item[]{\bf KG with Pairwise Sampling and Bias Functions ($\PWKG$-bias):} A GP with both offsets and bias functions is fitted,
    $\sigma_b^2, \eta^2, \sigma^2_w\geq 0$. Acquisition is the same as above.
    \item[]{\bf KG for Common Random Numbers with Compound Sphericity ($\CRNKG$-CS):} A GP with $\sigma_b^2=0$ and 
    $\eta^2,\sigma^2_w\geq 0$ is fitted. Acquisition can sample any seed according to $\CRNKG$.
    \item[]{\bf KG for Common Random Numbers ($\CRNKG$):} A GP with both offsets and bias functions
    is fitted, $\sigma_b^2, \eta^2, \sigma^2_w\geq 0$.  Acquisition can sample any seed according to $\CRNKG$.
    %
\end{itemize}

\subsection{Synthetic Data, no Bias Functions}

We set $X=\{1,..,100\}$ and generate synthetic data from a multivariate Gaussian 
$\tb(X) \sim N(\underline{0}, \ktb(X,X))$ where 
$\ktb(x,x') = 100^2\exp\left(-\frac{(x-x')^2}{2\cdot 5^2}\right)$. The offsets are sampled 
$o_s\sim N(0, \rho 50^2)$ and the white noise $w_s(x)\sim N(0, (1-\rho)50^2)$. 
We vary $\rho \in \{0,0.1,...,0.9,1.0\}$ holding the total noise constant such that standard KG will 
always perform the same. For algorithms we compare normal KG, $\PWKG$ and $\CRNKG$ all without bias functions.
For each method we evaluate the KG by Equation \ref{eq:KG_discrete} and set $A=X$. 
We optimize the acquisition function by exhaustive
search. In all cases we fit the GP
regression model with known kernel hyperparameters except for KG where we force $\rho=0$. This allows us to fully focus on differences in the generative model and acquisition function. We measure
opportunity cost, let $x_r^n=\argmax_{x}\mu^n(x,0)$,
\begin{equation}
\text{Opportunity Cost at time }n = \max \tb(x) - \tb(x_r^n).
\end{equation}
We report the frequency of seed reuse, how often at an iteration $n$ the next sampled
seed $s\n1$ was in the current history of observed seeds $S^n$.
If $\PWKG$ samples a pair for every iteration, the first sample of each pair
would be new and the second would be old hence the average reuse frequency
is upper bounded by $0.5$.

\begin{figure}[h]
\begin{center}
\includegraphics[width=0.8\textwidth]{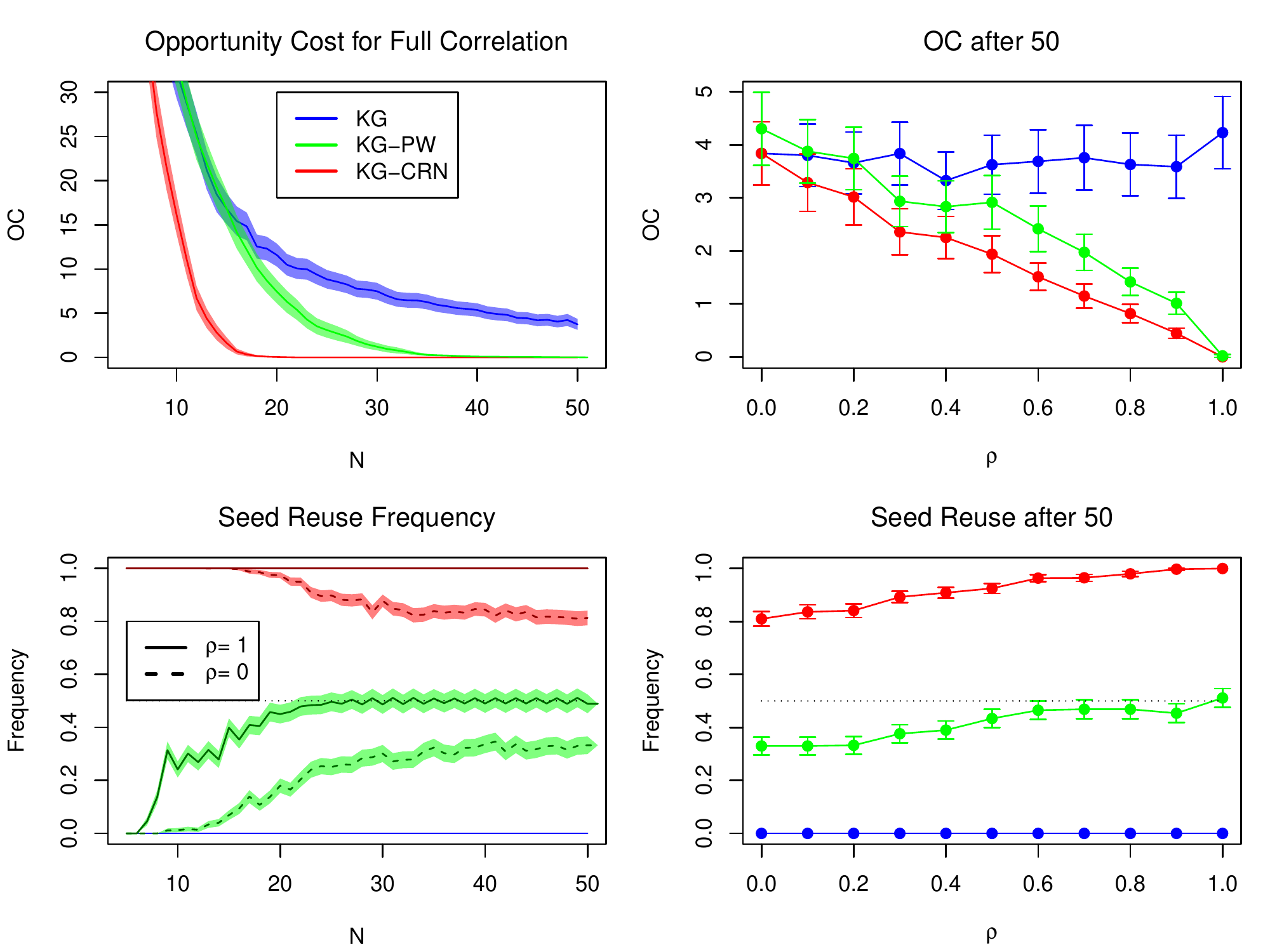}
\end{center}
\caption{
(top left) Opportunity Cost for the $\rho=1$ case, the $\rho=0$ case all algorithms equal KG. $\CRNKG$ aggressively optimizes a single seed.
(top right) final OC for a range of $\rho$ values. For increasing $\rho$ both CRN methods improve.
(bottom left) the average seed reuse for the cases $\rho=0,1$. For large $\rho$, $\PWKG$ is upper bounded by 0.5, $\CRNKG$ never samples a new seed.
(bottom right) final seed reuse over a range of $\rho$. 
\label{fig:vary_off}
}
\end{figure}

From top row plots of Figure~\ref{fig:vary_off}, for low $\rho$ values, all algorithms have similar
opportunity cost as there is no exploitable CRN structure. As $\rho$ increases
there is more CRN structure to exploit and $\PWKG$ performance improves for
larger budgets while $\CRNKG$ performance improves for all budgets.

The bottom row plots of Figure \ref{fig:vary_off} show seed reuse which we interpret
as how much an algorithm uses CRN.
For all $\rho$, $\CRNKG$ starts by resampling old seeds, utilizing CRN, and later
samples more new seeds only for low $\rho$, seed reuse dropping to 0.8, or querying
new seeds 20\% of the time. We see that this results in significantly faster convergence
in the $\rho=1$ case plotted.

$\PWKG$ instead starts by sampling singles on new seeds, ignoring CRN reproducing KG.
For larger budgets $\PWKG$ uses more pairs and improves upon KG for the range of $\rho$.
However for the best case for CRN, $\rho=1$, $\PWKG$ quickly hits its
seed reuse upper bound of 0.5, querying new seeds 50\% of the time, and cannot fully utilize CRN.

In the Electronic Companion~\ref{fig:EC_wiggle_experiments}, we present the same experiment
using only bias functions, 
and observe no improvement over standard KG, suggesting  that local 
differences correlation is not as beneficial as global, i.e. constant, correlation. 


\subsection{Assemble to Order Benchmark}
The Assemble to Order (ATO) simulator was introduced by \cite{xu2010industrial} and a slightly modified 
version has been used in \cite{xie2016bayesian} to test the $\PWKG$ algorithm.
A shop sells five products assembled from eight items held in inventory. A random
stream of customers arrives into the shop, each buying a product and consuming inventory.
When an item in inventory drops below a user defined threshold, an order for more is placed. The shop aims to
maximize profit, product sales minus storage cost, by optimizing the reorder thresholds for each item. A seed
defines the stream of customers and the item delivery times. For this problem, the solution space is $X=\{1,..,20\}^8$.

$\CRNKG_n(x,s)$ is evaluated and optimized as described in Section~\ref{sec:implementation}.
The expectation of the maximizations within $\PWKG(x\n1, x^{n+2})$ is evaluated exactly the same way and the function is optimized in two ways.
First, $x\n1$ is found using $\CRNKG$ on the new seed. $\PWKG(x\n1, x^{n+2})$
is then optimized over $X$ for $x^{n+2}$ only with the same multi-start gradient ascent optimizer. Second, including the best pair so far as one start, we  use multi-start gradient ascent over the full $X\times X$.

\begin{figure}[h]
\begin{center}
\includegraphics[width=0.8\textwidth]{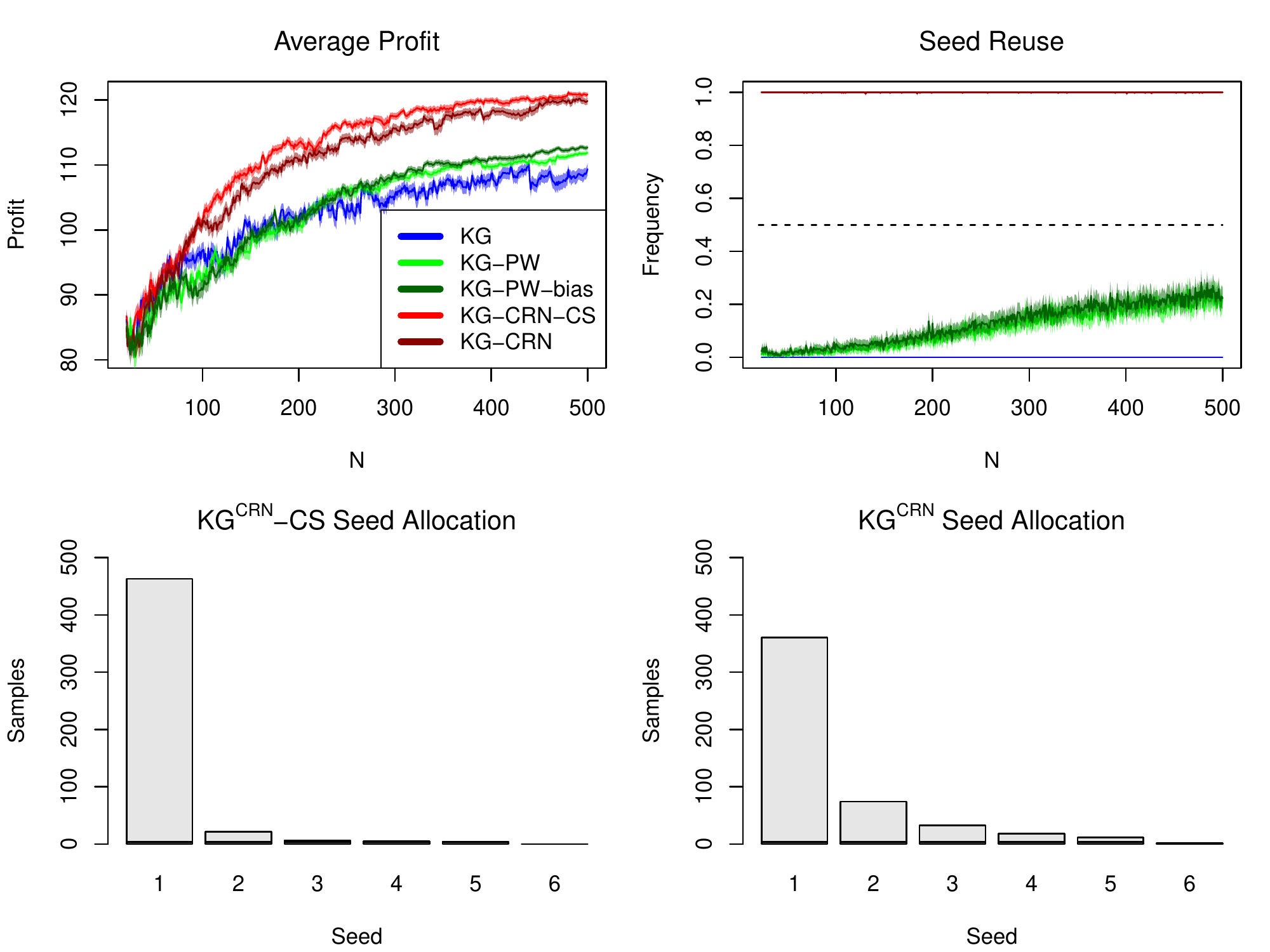}
\end{center}
\caption{Top left: profit of $x_r^N$ evaluated on a held-out set of 2,000 test seeds. 
Top right: average seed reuse over iterations.
Bottom: seed allocation for $\CRNKG$ without bias functions (left) and with bias functions (right).
Both $\CRNKG$ variants mostly sample a single seed.
\label{fig:ATO}
}
\end{figure}

All methods start with $n_{init}=20$. All hyperparameters are learnt by
maximum likelihood and fine tuned after each new sample.
We record the quality of the recommended $x_r^n = \amax{x}\mu^n(x,0)$
on a held-out test set of seeds.
ATO results are reported in Figure~\ref{fig:ATO}.

Both algorithms with $\CRNKG$ acquisition yield the largest profits
and the $\PWKG$ variants marginally improve upon KG. 
In this application, the $\CRNKG$ variants \textit{never} use new seeds
after the initial five seeds, instead allocating almost all budget to a
single seed suggesting that this ATO problem strongly benefits from reuse of seeds. From the previous experiment
we observed that $\CRNKG$ samples old seeds early and moves onto
new seeds for large budgets. In this learnt hyperparameter case,
as reported in the Electronic Companion EC.1, the offset hyperparameter, $\eta^2$, grows over
time as model fit improves and data collection focuses on the peak. 
Consequently, for larger budgets $\CRNKG$ is even more likely to
resample old seeds. With $\PWKG$, the early behavior samples singles (as opposed to pairs)
on new seeds which cannot inform any CRN hyperparameters and 
the algorithm never learns a larger offset parameter. As a result it allocates very little 
of the budget to pairs failing to significantly exploit the CRN structure and hence producing
marginally superior results to KG. In this application, the ability to revisit old seeds
clusters observations on fewer seeds which allows for more robust learning of CRN hyperparameters.




\subsection{Ambulances in a Square Problem}
\begin{figure}[h]
\begin{center}
\includegraphics[width=0.8\textwidth]{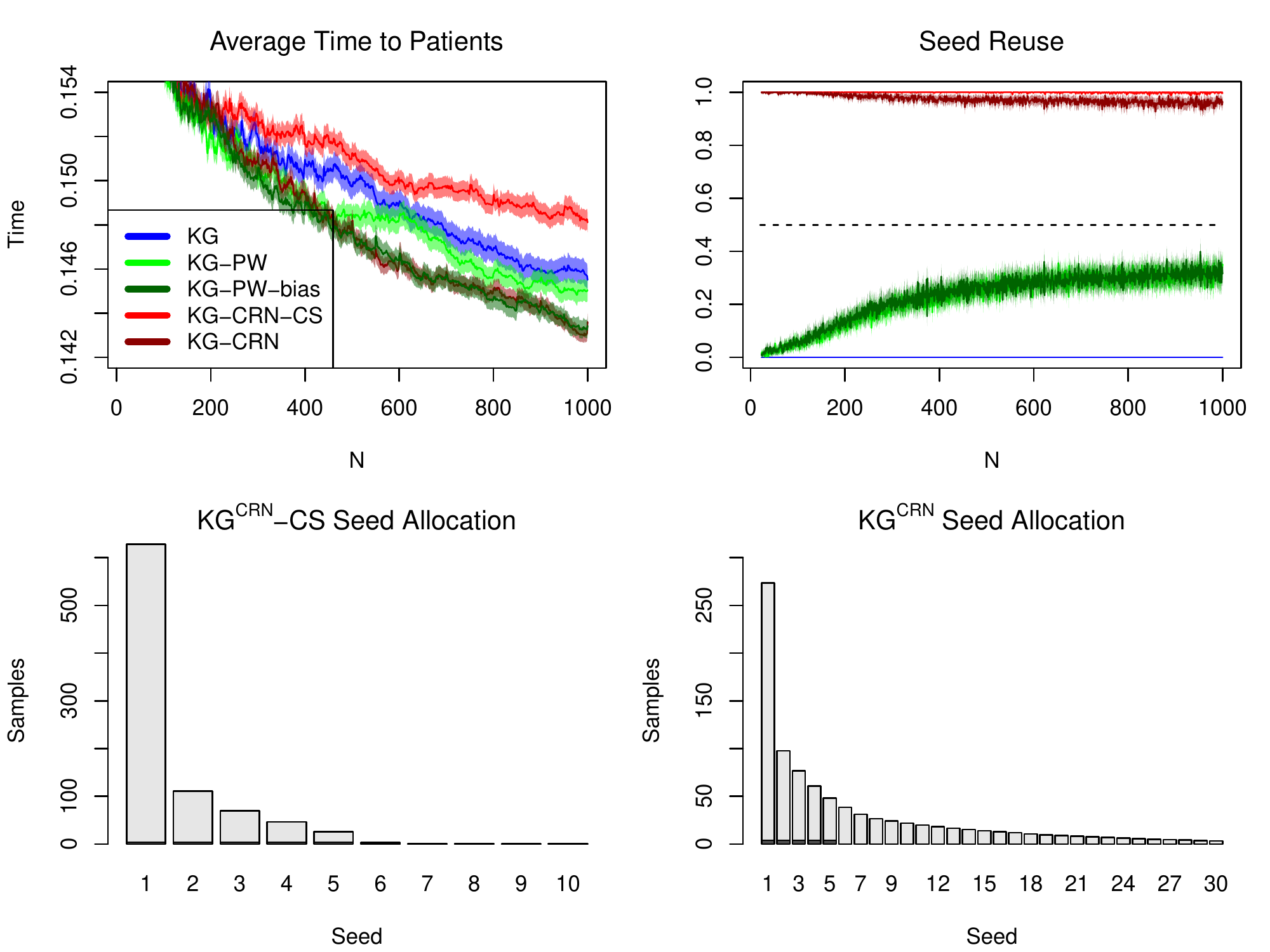}
\end{center}
\caption{
Top left: average journey time to patients. 
Top right: seed reuse over iterations.
Bottom: seed allocation by $\CRNKG$ without (left) and with (right) bias functions. 
The algorithms with bias functions
provide the best results. The compound spheric assumption, which is violated in this benchmark,
leads to greedy sampling of observed seeds and sub optimal performance.\label{fig:ambulances}}
\end{figure}

This simulator (AIS) was introduced by \cite{pasupathy2006testbed}.
Given a city over a 30km  by 30km square, one must optimize the
location of three ambulance bases to reduce the journey
time to patients that appear across the city as a Poisson
point process. The seed defines the times and locations of patients.
The solution space is $X=[0,30]^6$, the valid (x,y) locations for each of
three ambulance bases. We run the simulator for 1800 simulated 
time units in which on average 30 patients appear. This problem is
over a continuous search space and the optimal result for each 
realization of patients is to place the ambulance bases near 
the patients. Hence the peak $x$ of one seed is not the same as the
average of seeds and bias functions are required. 
Results are summarized in Figure~\ref{fig:ambulances}

Both algorithms with the surrogate model that includes bias functions provide the best results in this benchmark, marginally improving upon KG.
The $\CRNKG-\text{CS}$ algorithm that has the compound sphericity assumption in a continuous search space leads to excessive sampling of observed seeds
agreeing with Lemma~\ref{lem:cs_infX} and the conjectured behaviour of $\CRNKG$
acquisition. Our proposed $\CRNKG$ with bias functions on the other hand
does not suffer and automatically queries many new seeds. Again, both 
$\PWKG$ variants sample far more seeds which is less penalized in this benchmark. 

We also performed experiments where the sum of ambulance journey times was optimized and where the number of patients was fixed. 
All results, including ATO, are summarized in Table~\ref{tab:results}. In all experiments, the $\CRNKG\text{-CS}$ without bias functions never sampled a new seed. In the Electronic Companion we also report running time of all experiments and in all cases KG was quickest, followed by the $\CRNKG$ variants and the $\PWKG$ variants used the most computational time.


\ifwithpics
\begin{table}[h]
    \centering
    
    \caption{\label{tab:results}Mean $\pm$ 2 standard errors of average performance for all benchmarks,
results that do not significantly differ from the best are in bold. The ability to revisit
seeds improves the ATO results and including bias functions improves AIS results
(or compound sphericity significantly harms AIS). 
}
\begin{small}

\begin{tabular}{|c|c|c|c|c|c|}
\hline 
                 & KG                &  $\PWKG$          & $\PWKG$-bias  &   $\CRNKG$-CS    & $\CRNKG$ \\
\hline
ATO, N=500       & $109.35\pm 1.88$  & $111.86\pm 0.65$    & $112.69\pm 0.67$  & \textbf{120.99 $\pm$ 0.71} & \textbf{119.84 $\pm$ 1.13} \\
AIS, N=500       & $.1498\pm .0011$  & \textbf{.1483 $\pm$ 0.0010} & \textbf{.1477 $\pm$ .0010} & $.1512\pm .0010$  & \textbf{.1482$ \pm$ .0010}\\
AIS, N=1000      & $.1455\pm .0010$  & $.1450 \pm 0.0010$ & \textbf{.1435 $\pm$ .0009} & $.1481\pm .0009$  & \textbf{.1436 $\pm$ .0008}\\
AIS, sum time    & $4.66 \pm 0.33$   & $4.611 \pm .045$   & \textbf{4.449$\pm$ .030}   & $4.515\pm .035$   & \textbf{4.430 $\pm$ .034}\\
AIS, 30 patients & $.1498\pm .0009$  & \textbf{.1468 $\pm$ .0008}  & \textbf{.1467 $\pm$ .0009} & .1482 $\pm$ .0008  & \textbf{.1467 $\pm$ .0009}\\
\hline
\end{tabular}
\end{small}

\end{table}
\fi

Therefore both the ability to revisit old seeds and the modelling of bias functions are necessary to make a robust algorithm that works across a variety of problems.

\section{Conclusion} \label{sec:conclusion}

We proposed a Bayesian approach to simulation optimization with common random numbers where the seed
of the random number generator used within a stochastic objective function is an input to 
be chosen by the optimization algorithm. We augment a standard Gaussian process model with 
two extra hyperparameters to model structured noise (seed/scenario influence), while maintaining the ability to predict the
average output of the target function in closed form. Matching this augmented model, we propose $\CRNKG$
that quantifies the benefit of evaluating the objective for a given solution and seed, providing a clean  framework that allows Bayesian optimization to automatically exploit CRN where this is beneficial, and recovers standard KG where not.
Moreover, the proposed $\CRNKG$ algorithm structure does not add significant computational burden
over the equivalent non-CRN Knowledge Gradient due to the fundamental structure of CRN.

In this work we focus on global optimisation, in future work we plan to augment other problem settings with common random numbers, such as multi-fidelity
optimization, simulations with input uncertainty, and multi-objective optimization.
%

\bibliographystyle{plain}
\bibliography{Sections/CRNKGbib}

\appendix

\section{Proofs of Statements}

%
%
%
%
\subsection{Estimating the Target}
The data collected and the surrogate model are over the domain $X\times \Np$ whereas
the target of optimization is a function over $X$ and we show how to derive an estimate
for the target.
This result is an immediate consequence of the symmetry
of the model across unobserved seeds proven in Lemma \ref{lem:equal_mu_kern}. 
As a result of this symmetry, when taking the limit over infinite seeds, unobserved
seeds dominate proving in Lemma~\ref{prop:thetabar_gp} yielding a simple form of the GP posterior for the target.
This result is consistent with other CRN and non-CRN methods that do not make
the seed explicit but do incorporate off-diagonal noise covariance matrix.

\begin{repeatlemma}[Lemma \ref{prop:thetabar_gp}]
For any given kernel over the domain $X\times \Np$ that is of the form $k_{\bar\theta}(x,x') + \delta_{ss'}k_\epsilon(x,x')$, 
and a dataset of $n$ input-output triplets $D^n$, the posterior over the target is a Gaussian process given by
\begin{eqnarray}
\bar\theta(x)|D^n  &\sim & GP(\mu^n_{\bar\theta}(x), k^n_{\bar\theta}(x,x')), \\
\mu^n_{\bar\theta}(x) &=&  \mu^n(x,s'), \\
k^n_{\bar\theta}(x, x') &=&  k^n(x,s', x', s''),
\end{eqnarray}
where $s', s''\in \Np\setminus S^n$ with $s'\neq s''$ any two unobserved unequal seeds.
\end{repeatlemma}

The Gaussian process model is over the domain $X\times \Np$, with infinite seeds. 
The following result states that the Gaussian process model makes
identical predictions for all the unobserved seeds.
\begin{lemma}\label{lem:equal_mu_kern}
Let $\theta(x,s)$ be a realization of a Gaussian Process with $\mu^0(x,s)=0$ and any positive 
semi-definite kernel of the form 
$k(x,s,x',s') = \ktb(x,x')+\delta_{ss'}k_\epsilon(x,x')$. For all $x\in X$, $s_{obs}\in S^n$, and 
unobserved seeds $s,s',s''\in \Np\setminus S^n$, the posterior mean and kernel satisfy
\begin{eqnarray}
\mu^n(x,s) &=& \mu^n(x,s'), \\
k^n(x,s_{obs},x',s) &=& k^n(x,s_{obs},x',s'),\label{eq:kern_old_new}\\
k^n(x,s,x',s') &=& k^n(x,s,x',s'') =  k^n(x,s',x',s'').\label{eq:kern_new_new}
\end{eqnarray}
\end{lemma}
\proof{Proof} Writing out the posterior mean in full from Equation \ref{eq:GPn},
\begin{eqnarray*}
\mu^n(x,s) &=& k^0(x,s,\tXn)K^{-1}Y^n \\
&=&
 \begin{cases} 
      \big(\ktb(x,X^n) + (\One_{s= S^n}\trans \circ k_\epsilon(x,X^n)) \big)K^{-1}Y^n & s \in S^n \\
      \ktb(x,X^n)K^{-1}Y^n      &    s \in \Np\setminus S^n 
   \end{cases}
\end{eqnarray*}
where $a\circ b$ is element-wise product $\One_{s= S^n}\in\{0,1\}^n$ is a binary masking column vector that is zeros
for all $s\in \Np\setminus S^n$. The proofs for 
Equations \ref{eq:kern_old_new} and \ref{eq:kern_new_new} follow similarly from Equation \ref{eq:GPk}.
\Halmos
\endproof

We next prove the main lemma. The target of optimization is the infinite average over seeds,
and the Gaussian process model makes identical predictions for unobserved seeds. The infinite average
is dominated by unobserved seeds with identical predictions. Hence we may simply use
the prediction of any one unobserved seed as a model for the infinite average/target.
\proof{Proof of Lemma~\ref{prop:thetabar_gp}}
The target of optimization, $\tb(x)$, is given by the average output over infinitely many 
seeds which may be written as the limit
\begin{eqnarray}
\tb(x) \,&=&\, \lim_{N_s \to \infty} \frac{1}{N_s}\sum_{s=1}^{N_s}\theta(x,s).
\end{eqnarray}
Adopting the shorthand $\E_n[...]=\E[...|D^n]$, we first consider the posterior expected performance,
\begin{eqnarray}
\E_n[\tb(x)] &=& \E_n\left[  \lim_{N_s \to \infty} \frac{1}{N_s}\sum_{s=1}^{N_s}\theta(x,s) \right] \label{eq:mu_infsumseeds}\\
&=& \lim_{N_s \to \infty} \frac{1}{N_s}\sum_{s=1}^{N_s} \E_n\left[  \theta(x,s)\right]\\
&=&\lim_{N_s \to \infty} \frac{1}{N_s}\sum_{s=1}^{N_s}  \mu^n(x,s). 
\end{eqnarray}

Let $n_s=\max\{S^n\}$ be the largest observed seed. The sum of posterior 
means can be split into sampled seeds $s \in \{1,...,n_s\}$ and unsampled seeds 
$s \in \{ n_s+1,....,N_s \}$,
\begin{eqnarray}
\E_n[\tb(x)]  &=&  \lim_{N_s \to \infty} \frac{1}{N_s}\left(\sum_{s=1}^{n_s} \mu^n(x,s) + \sum_{s'=n_s+1}^{N_s} \mu^n(x,s') \right) \\
&=&  \lim_{N_s \to \infty} \frac{1}{N_s}\left(\sum_{s=1}^{n_s} \mu^n(x,s) + (N_s-n_s)\mu^n(x,n_s+1) \right) \\
&=&  \lim_{N_s \to \infty} \frac{1}{N_s}\left(\sum_{s=1}^{n_s} \mu^n(x,s)-n_s\mu^n(x,n_s+1) \right)+ \mu^n(x,n_s+1)\\
&=& \mu^n(x,n_s+1),
\end{eqnarray}
where we have used Lemma \ref{lem:equal_mu_kern} to simplify. 
Similarly for the covariance, writing each $\tb(x)$ term as the limit of a sum over seeds,
\begin{eqnarray}\label{eq:kernel_sum}
&&\E_n \bigg[ \big(\tb(x)-\E_n[\tb(x)]\big) \big(\tb(x')-\E_n[\tb(x')] \big) \bigg] \\
&=& \E_n\left[ \left( \lim_{N_s\to \infty}\frac{1}{N_s}\sum_{s=1}^{N_s}\theta(x,s)-\mu(x,s) \right) \left( \lim_{N_t\to \infty}\frac{1}{N_t}\sum_{s'=1}^{N_t}\theta(x',s')-\mu(x',s')         \right)\right]\\
&=& \lim_{N_s, N_t\to \infty}\frac{1}{N_sN_t}\sum_{s, s'=1}^{N_s,N_t}\E_n\left[ \left( \theta(x,s)-\mu(x,s)         \right)\left( \theta(x',s')-\mu(x',s')         \right)\right]\\
&=& \lim_{N_s, N_t\to \infty}\frac{1}{N_sN_t}\sum_{s, s'=1}^{N_s,N_t}k^n(x,s,x's'). \label{eq:kern_lim_sum}
\end{eqnarray} 
The domain in the limit of the summation, $\Np\times\Np$, is unaffected by setting $N_t=N_s$. The summation decomposes into four terms,
\begin{eqnarray*}
\sum_{s,s'=1}^{N_s} k^n(x,s, x',s') 
&=&  \underbrace{\sum_{s,s'=1}^{n_s}k^n(x,s,x',s') }_{\text{observed seeds full covariance}}  +\underbrace{\sum_{s'=n_s+1}^{N_s}\sum_{s=1}^{n_s}k^n(x,s,x',s')}_{\text{observed-unobserved covariance}}\\
&& + \underbrace{\sum_{s=n_s+1}^{N_s}k^n(x,s,x',s)}_{\text{unobserved seeds variance}} 
+\underbrace{\sum_{n_s < s\neq s' \leq N_s}k^n(x,s,x',s')}_{\text{unobserved seeds covariance}} \nonumber \\ \\
%
%
%
&=& \underbrace{\sum_{s,s'=1}^{n_s}k^n(x,s,x',s')}_{\text{constant with $N_s$}}  
+\underbrace{ 2(N_s-n_s)\sum_{s=1}^{n_s}k^n(x,s,x',s')}_{\text{linear with $N_s$}}\\
&& + \underbrace{(N_s-n_s) k^n(x, s',x',s')}_{\text{linear with $N_s$}}
+ \underbrace{(N_s-n_s)^2 k^n(x, s',x',s''),}_{\text{quadratic with $N_s$}} \nonumber
\end{eqnarray*}
where $s'$ and $s''$ are two unequal unobserved seeds. Dividing the final Equation by $N_s^2$ and taking the 
limit $N_s\to\infty$, only the final term remains. 
\Halmos
\endproof

Given the assumed kernel with independent and identically distributed difference functions,
the average of infinitely many seeds includes finite observed seeds and 
infinitely many identical unobserved seeds and unobserved seeds. Unobserved seeds dominate the infinite average and the performance under any unobserved seed 
is an estimator for the objective function.
Likewise the posterior covariance between infinite averages is the posterior covariance
between any two unique unobserved seeds.
Also note that the prior kernel for the objective evaluated at different seeds returns the
prior kernel for the target $\bar k^0(x,x')=k^0(x,1,x',2)=k_{\tb}(x,x')$ as desired.

%
%
%
\subsection{Proof of Theorem \ref{thm:KG_assymptotic}}
We next show that, under certain assumptions on the target function,
given an infinite sampling budget, $N\to \infty$, the $\CRNKG$ algorithm
will discover the true optimum. We first restate the result.

\begin{repeattheorem}[Theorem \ref{thm:KG_assymptotic}]
Let $x^N_r\in A$ be the point that $\CRNKG$ recommends in iteration $N$. For each $p\in[0, 1)$ 
there is a constant $K_p$ such that with probability $p$
$$ \lim_{N\to \infty}\tb(x^N_r) > \tb(x^{OPT}) - K_pd. $$
\end{repeattheorem}
We first prove properties of the $\CRNKG_n(x,s)$ function and then consider the error due to discretization.

Lemma~\ref{lem:inf_exists} ensures the GP model exists in the limit of infinite data. We then show that $\CRNKG_n(x,s)$ is non-negative in Lemma~\ref{lem:KGpositive} and that it is zero for sampled input pairs in Lemma~\ref{lem:zeroKG}. We then show that if a single $x$ is sampled for infinitely many (not necessarily consecutive) seeds, again $\CRNKG_n(x,s)$ tends to zero also for all unevaluated seeds in Lemma~\ref{lem:zero_post_truth}. Then in Lemma~\ref{lem:zeroKG_constSIGT} we show the opposite direction, if $\CRNKG_n(x,s)$ is zero, this implies that the peak of the target prediction will not change by sampling $(x,s)$. This is extended in Lemma~\ref{lem:all_zeroKG_known_max} that states that if for a new seed $s$, $\CRNKG_n(x,s)=0$ for all $X$ then no more samples will change the peak prediction of the target and the the true peak is known
when $X$ is a discrete set.

The error due to discretization relies on the assumption of a differentiable GP kernel, such as Mat\'ern, and
a using a Lipschitz continuity argument, the error may be bounded proving Theorem~\ref{thm:KG_assymptotic}.
%
%
%
%
%
%
%
%
%
%
%
%
%
%
The following result simply states that the GP model exists in the limit of infinite data.
First we define $V^n(x,x')=\E_n[\tb(x)\tb(x')]$.
\begin{lemma} \label{lem:inf_exists}
Let $x,x'\in X$. Then the limits of the series $(\mub^n(x))_n$ and $(V^n(x,x'))_n$ exist and are denoted by $\mub^\infty(x)$ and $V^\infty(x,x')$, respectively. Then we have 
\begin{eqnarray}
\lim_{n\to \infty} \mub^n(x) &=& \mub^\infty(x) \\
\lim_{n\to \infty} V^n(x,x') &=& V^\infty(x,x')
\end{eqnarray}
almost surely. 
\end{lemma}
\proof{Proof} $\tb(x)$ and $\tb(x)\tb(x')$ are integrable random variables for all $x,x'\in X$ by choice 
of $\tb$. Proposition 2.7 in \cite{ccinlar2011probability} states that any sequence of conditional expectations 
of an integrable random variable under an increasing filtration is uniformly integrable martingale. Thus, both 
sequences converge almost surely to their respective limit. 
\Halmos
\endproof

%
%
%
%
%
%
%
%
%
%
%
%
The next result states the $\CRNKG(x,s)$ is non-negative for all input pairs.
\begin{lemma} \label{lem:KGpositive}
$\CRNKG_n(x,s)\geq 0$ holds for all $(x,s)\in X\times \Np$.
\end{lemma}
\proof{Proof} Adopting the shorthand $x^n_r = \amax{x\in X} \mu^n(x,0)$, we may write 
$\max_x \mu^n(x, 0) = \mu^n(x_r^n, 0)$ and
\begin{eqnarray*}
\CRNKG_n(x,s) &=& \E\left[ \max_{x'\in X} \mu^n(x',0) +\st^n(x',0; x,s)Z - \mu^n(x^n_r,0) \right]  \\
&=& \E\left[ \max_{x'\in X} \mu^n(x',0) +(\st^n(x',0; x,s) - c)Z - \mu^n(x^n_r,0) \right]
\end{eqnarray*}
where the expectation is over $Z\sim N(0,1)$ and $c$ is an arbitrary constant.
In particular, by setting $c = \st^n(x^n_r,0; x,s)$, the inner expression,
when evaluated at $x_r^n\in X$, satisfies 
$$\mu^n(x_r^n,0) +(\st^n(x_r^n,0; x,s) - \st^n(x_r^n,0; x,s))Z - \mu^n(x^n_r,0) = 0$$
for all $Z\in  \R$ and
\begin{eqnarray*} \label{eq:kg_positive}
\max_{x'\in X}\{ \mu(x',0) +(\st^n(x',0; x,s) - c)Z -\mu_0\} 
\,\,\geq\,\, \mu(x^n_r,0) +(\st^n(x^n_r,0; x,s) - c)Z - \mu^n(x^n_r,0)
\,\,=\,\,0
\end{eqnarray*}
for all $Z$ and $\CRNKG_n(x,s)$ may be written as the expectation of a non-negative random variable.
\Halmos
\endproof

%
%
%
%
%
%
%
%
The following result states that once an input pair $(x,s)$ has been observed, $\theta(x,s)$ is 
known and $\CRNKG_n(x,s)$ is zero. Combined with the result that $\CRNKG_n(x,s)$ is non-negative, it follows
that observed input pairs $(x,s)$ are minima of the function $\CRNKG_n(x,s)$.
\begin{lemma}\label{lem:zeroKG}
Given deterministic simulation outputs, there is no improvement in re-sampling a sampled point.
$$\CRNKG_n(x^i,s^i) = 0$$ 
for all $(x^i,s^i)\in \tXn$.
\end{lemma} 
\proof{Proof} The posterior covariance between the output at any point and the output at an observed point is zero,
writing out the full matrix multiplication for the posterior kernel and simplifying yields
\begin{eqnarray*}
k^n(x^i,s^i; x,s) &=& k^0(x^i,s^i; x,s) - k^0(x^i,s^i; \tXn)\left(  k^0(\tXn; \tXn) \right)^{-1} k^0(\tXn; x, s) \\
&=& k^0(x^i,s^i; x,s) - \left[k^0(\tXn; \tXn)\right]_i \left(  k^0(\tXn; \tXn) \right)^{-1} k^0(\tXn; x, s) \\
&=& k^0(x^i,s^i; x,s) - {\One^n_i}\trans k^0(\tXn; x, s) \\
&=& k^0(x^i,s^i; x,s) -  k^0(x^i,s^i; x, s)\\
&=&0
\end{eqnarray*}
where $[\cdot]_i$ is the $i^{th}$ row. The second line contains the $i^{th}$ row of a matrix multiplied
by its inverse returning the $i^{th}$ row of the identity matrix denoted ${\One^n_i}\trans$.
Therefore $\st^n(x,s;x^i,s^i)=0$ for all $(x,s)$ and $\CRNKG_n(x,s) =0$.
\Halmos
\endproof

%
%
%
%
%
%
%
%
Let $\omega$ denote an arbitrary sample path, $\omega =((x,s)^1,(x,s)^2,......)$, determining
an input pair for each query to the objective as $n\to \infty$. 
Lemmas \ref{lem:KGpositive} and \ref{lem:zeroKG} imply sampled point inputs are minima
of $\CRNKG$ and recall that according to the algorithm, new samples are allocated to 
maxima $(x,s)\n1 = \amax{} \CRNKG_n(x,s)$.
These facts together imply that no input $(x,s)$ will be sampled more than once. We need only to 
consider sample paths $\omega$ where all sampled inputs pairs $(x^i,s^i)$ are unique. Recall that we suppose a 
(finite) discretization of~$X$, thus there must be an 
$x\in X$ that is observed for an infinite number of seeds on $\omega$ as~$n \to \infty$. 
We study the asymptotic behaviour $\CRNKG_n(x,s)$ for $n\to \infty$ 
as a function of $\mu^n(x,0)$, $\st^n(x',0,x,s)$.

%
%
%
%
If $s$ is an new seed and $x$ has been observed for infinitely many seeds, the next result states that $\CRNKG_n(x,s)$ tends to zero, there is less/no value in re-evaluating $x$ for another new seed.
\begin{lemma} \label{lem:zero_post_truth}
If $x$ is sampled for infinitely many (not necessarily consecutive) seeds,
then $\st^\infty(x',0;x,s)=0$ for all $x'\in X$ and all $s\in \Np$ and 
$\CRNKG_\infty(x,s)= 0$ for all $s\in\Np$ almost surely.
\end{lemma}
\proof{Proof} 
Setting $x\n1=x$ and assuming $(x^i,s^i)$ pairs are arranged such that $s\n1$ is always a new seed, the 
posterior variance reduces to zero
\begin{eqnarray*}
\lim_{n\to\infty} |\st^n(x',0;x,s\n1)| &=& \lim_{n\to\infty} \frac{|k^n(x',0,x,s\n1)|}{\sqrt{k^n(x,s\n1,x,s\n1)}} \\
&=& \lim_{n\to\infty} \frac{\bar k^n(x',x)}{\sqrt{\bar k^n(x,x)+k_\epsilon(x,x)}} \\
&\leq& \lim_{n\to\infty} \sqrt{\bar k^n(x',x')} \frac{\sqrt{\bar k^n(x,x)}}{\sqrt{\bar k^n(x,x)+k_\epsilon(x,x)}} \\
&=& 0
\end{eqnarray*}
where the final line is by noting that $\bar k^n(x,x)+k_\epsilon(x,x)>0$ for all $n$ and $x$.
\Halmos
\endproof

%
%
%
%
%
%
The following result states that if there is no benefit of a new measurement for an input pair $(x,s)$,
then the change in the posterior mean, $\st^n(x',0; x,s)$ must be constant, i.e. the new sample at $(x,s)$
will only have the effect of adding a constant to the prediction of the target, hence learning nothing about the 
peak of the target. The contrapositive is that for input points for which $\st^n(x',0; x,s)$ varies with $x'$, $\CRNKG$ is strictly positive.
\begin{lemma} \label{lem:zeroKG_constSIGT}
Let $(x,s)$ be an input pair for which $\CRNKG_n(x,s)= 0$. Then for all $x'\in X$
$$ \st^n(x',0; x,s) = c$$
where $c$ is a constant.
\end{lemma}
\proof{Proof} From Equation \ref{eq:kg_positive}, $\CRNKG_n(x,s)$ can be written as the expectation of 
a non-negative random variable. Therefore the random variable itself must equate to zero almost surely implying
\begin{eqnarray*}
\max_{x'\in X}\{ \mu(x',0) +(\st(x',0; x,s) - c)Z - \mu^n(x^n_r,0)\} &=&0 \\
\max_{x'\in X}\{ \mu(x',0) +(\st(x',0; x,s) - c)Z \} &=& \max_{x''\in X}\{ \mu(x'',0)\}
\end{eqnarray*}
for all $Z\in \R$. This implies $\st(x',0; x,s)= c$ for all $x'\in X$.
\Halmos
\endproof

Note the case where $(x,s)\in\tXn$ we have that $\st^n(x',0; x,s)=0$ for all $x'\in X$.

%
%
%
%
%
We next show that is there is no value in evaluating any input pair, then 
the optimizer of the target is known.
\begin{lemma} \label{lem:all_zeroKG_known_max}
Let $s\in \Np \setminus S^n$ be an unobserved seed, if $\CRNKG_n(x,s)=0$
for all $x\in X$, then $\amax{x} \mu^n(x,0) = \amax{x}\tb(x)$
\end{lemma}
\proof{Proof} By Lemma \ref{lem:zeroKG_constSIGT}, we have that $\bar k^n(x,x')=c$ for all $x, x'\in X$
and the covariance matrix $\bar k^n(X,X)$ is proportional to the all ones matrix. Hence $\bar\theta(x)-\mu^n(x,0)$
is a normal random variable that is constant across all $x\in X$ and $\amax{x\in X}\mu(x,0)=\amax{x\in X}\tb(x)$ holds.
\Halmos
\endproof

%
%
%
%
%
%

Lemmas \ref{lem:zeroKG}, \ref{lem:zero_post_truth}, consider evaluating $\CRNKG$ as 
the sampling budget increases in a specific way. More generally, recall that $\CRNKG$ picks $(x,s)\n1\in \amax{} \CRNKG_n(x,s)$ in each iteration $n$. Since $\theta(x,\cdot)$ is evaluated infinitely often (by choice of $x$), 
$\CRNKG_n(x,\cdot)\to 0$ for all $x\in A$ holds almost surely and by Lemma \ref{lem:all_zeroKG_known_max}
the true optimizer is known. 
%

%
%
%
%
%
%

Next, we consider a bound on the loss due to discretization of a continuous search space. 
Suppose that $X\subset \R^d$ is a compact infinite set and $A\subset X$ is a finite set 
of discretization points. Suppose that $\bar{\mu}^0(x)=0$ for all $x$, and $k_{\tb}(x, x')$ 
is a four times differentiable Mat\'ern kernel e.g. the popular squared exponential kernel. 
Suppose that $\tb(x)$ is drawn from the prior, i.e. let $\tb(x)\sim \GP(\bar{\mu}^0(x), k_{\tb}(x, x'))$ 
then the sample $\tb(x)$ over the set of functions is itself twice differentiable in $X$ with probability one. 
Let $x^{OPT}=\amax{x\in X}{\tb(x)}$ and $d=\max_{x'\in X}\min_{x\in A}\text{dist}(x,x')$ be the 
largest distance from any point in the continuous domain $X$ to it's nearest neighbor in $A$.

\proof{Proof}
The extrema of $\frac{\delta}{\delta x_i} \tb(x)$ over $X$ are bounded, the partial derivatives of $\tb(x)$ 
are also GPs for our choice of $k_{\tb}^0(x,x)$. Thus we can compute for every $p\in [0,1)$ a constant $K_p$ such 
that $\tb(x)$ is $K_p$ Lipschitz continuous on $X$ with probability at least $p$, then there exists an 
$\bar x \in A$ with $\text{dist}(\bar x, x^{OPT})\leq d$ and
$$\tb(\bar x) > \tb(x^{OPT})-K_pd$$
holds with probability $p$. Finally the point recommended by $\CRNKG$ is the 
maximizer of $x^N_r\in\amax{x\in A}\tb(x)$ and therefore is not worse than $\bar x$
\begin{eqnarray*}
\lim_{N\to\infty}\tb(x^N_r) &\geq& \tb(\bar x) \\
&\geq& \tb(x^{OPT})-K_pd
\end{eqnarray*}
\Halmos
\endproof

Thus when applying the $\CRNKG$ algortihm to a disctretized search space, the true optimizer becomes known
is the sampling budget increases without bound and if the underlying target function is continuous,
the error is bounded simply due to Lipschitz continuity.

%
%
%
%
%

%
%
%
\subsection{Proof of Lemmata \ref{thm:none_left} and \ref{prop:KGCS_never_new}}
%
%
We next provide proofs for the $\CRNKG$ algorithm behaviour in the case of 
compound sphericity with full noise correlation, recall this corresponds
to the difference functions reducing to constant offsets and an algorithm may 
optimize one seed as a single seed is a deterministic function with the same optimizer as the target.
This is essentially a best-case scenario for optimization with common random numbers.
Lemma \ref{lem:cs_cor} of the main paper states that the difference $\mu^n(x,s)-\mu^n(x,s')=A_s - A_{s'}$ is 
constant for all $x$. Likewise the same relationship applies to $\st^n(x',s';x,s)$ that 
quantifies changes in the posterior mean and therefore must also maintain the 
symmetry over seeds $s'$. All results in this section assume $\theta(x,s)$ is a realization of a Gaussian process
with the compound spheric kernel and full correlation $k_\epsilon(x,x')=\eta^2$.

This first result states that, when sampling a point $(x,s)$ the update in
the prediction for one seed differs from the update in prediction for another seed by an 
additive constant. Predictions for all seeds have the same shape/gradient and differ
only by global constants.
\begin{lemma} \label{lem:sigt_all_diff_constant}
Let $x,x'\in X$, $s,s'\in \Np$, then the difference in posterior mean updates satisfies
$$\st^n(x',s';x,s) = \st^n(x',0;x,s) + h^n(s',x,s).$$
\end{lemma}
\proof{Proof}
\begin{eqnarray*}
\st^n(x',s';x,s) 
&=& \dfrac{ k^n(x',s'; x,s) }{ \sqrt{k^n(x,s,x,s)} } \\
&=& \dfrac{1}{\sqrt{k^n(x,s,x,s)}}
\left(
\ktb(x',x) + \eta^2\delta_{ss'} -\bigg(\ktb(x', X^n)  + \eta^2\One_{s'=S^n}\trans\bigg) K^{-1} \bigg(\ktb(X^n, x) + \eta^2 \One_{s=S^n}   \bigg)
\right) \\
&=& \st^n(x',0;x,s) + \underbrace{\dfrac{
\eta^2\delta_{ss'} -{\eta^2\One_{s'=S^n}\trans} K^{-1} \bigg(\ktb(X^n, x) + \eta^2\One_{s=S^n} \bigg)
}{
\sqrt{k^n(x,s,x,s)}
}}_{\text{independent of $x'$}} \\
&=& \st^n(x',0;x,s) +  h^n(s',x,s)
\end{eqnarray*}
\Halmos
\endproof

%
%
%
%
%
%

As a result of the symmetry over seeds it is possible to use any seed $s\in \Np$ as the 
target of optimization formalized in the following Lemma.
\begin{lemma} \label{lem:KG_add_invariant}
Let $x\in X$, $s,s'\in \Np$, then
$$\CRNKG_n(x,s) = \E[\max_{x'\in X}\mu^n(x',s') + \st^n(x',s';x,s)Z - \max_{x''\in X}\mu^n(x'',s')].$$
\end{lemma}
\textit{Proof}
\begin{eqnarray*}
\CRNKG_n(x,s) &=& \E[\max_{x'\in X}\mu^n(x',0) + \st^n(x',0;x,s)Z - \max_{x''\in X}\mu^n(x'',0)] \\
&=& \E[\max_{x'\in X}\mu^n(x',s') - A_{s'} + (\st^n(x',s';x,s)-h(s',x,s))Z - \max_{x''\in X}\mu^n(x'',s')-A_{s'}] \\
&=& \E[\max_{x'\in X}\mu^n(x',s') + \st^n(x',s';x,s)Z - \max_{x''\in X}\mu^n(x'',s')] -h(s',x,s)\E[Z]  \\
&=& \E[\max_{x'\in X}\mu^n(x',s') + \st^n(x',s';x,s)Z - \max_{x''\in X}\mu^n(x'',s')]
\end{eqnarray*}
%
%
%
%
%
%
%
We next prove Lemma~\ref{thm:none_left} from the main paper: if there are finite solutions $X$ 
and all have been evaluated on a common seed, then there is no more value of sampling any solution on any seed.
\begin{repeatlemma}[Lemma \ref{thm:none_left}]
Let $X=\{x_1,...,x_d\}$ and $\tXn=\{(x_1, 1),...,(x_d, 1)\}$  then for all $(x,s)\in X\times \Np$
\begin{equation*}
\CRNKG_n(x,s)=0
\end{equation*}
and the maximizer $\argmax{x}\tb(x)$ is known.
\end{repeatlemma}

\proof{Proof}
Lemma \ref{lem:KG_add_invariant} shows that any seed can be used as the target of optimization.
Therefore we may choose $s=1$ as the target. All $x$ have been sampled for $s=1$ therefore 
$\st^n(x,1;x',s')=0$ for all $x\in X$ and $s'\in \Np$. Hence 
\begin{eqnarray*}
\CRNKG(x,s) &=& \E[\max_{x'\in X}\mu^n(x',1) + 0Z - \max_{x''\in X}\mu^n(x'',1)]\\
&=& 0
\end{eqnarray*}
for all $x,s \in X \times \Np$. By Lemma \ref{lem:all_zeroKG_known_max} the maximizer 
$\amax{x \in X}\tb(x)$ is known (although it's underlying value, $\max\tb(x)$, is not known).
\Halmos
\endproof
%
%
%
%
%

%
%
%
%
%
We next prove the result from the main paper that $\CRNKG$, when evaluated as in \cite{KGCP_scott2011correlated},
never samples a new seed and reduces to Expected Improvement (EI) of \cite{jones1998efficient}.
\begin{repeatlemma}[Lemma \ref{prop:KGCS_never_new}]
Let $X\subset \R^d$ be a set of possible solutions, $\tXn=\{(x^1,1),...,(x^n,1)\}$ be the set of sampled
input pairs and $X^n=(x^1,...,x^n)$. Define
\begin{eqnarray*}
\CRNKG_n(x,s; A) = \E\bigg[\max_{x'\in A\cup \{x\}}\mu\n1(x',0)- \max_{x'\in A\cup \{x\}} \mu^n(x',0)\bigg| D^n, (x,s)\n1=(x,s)\bigg].
\end{eqnarray*}
Then for all $x \in X$
$$\CRNKG_n(x,1; X^n)>\CRNKG_n(x,2; X^n)$$
and therefore $\max_x\CRNKG_n(x,1; X^n)>\max_x\CRNKG_n(x,2; X^n)$ and
seed $s=2$ will never be evaluated.
Further 
$$\CRNKG_n(x,1; X^n) = \E\big[\max\{ 0, y\n1 - \max Y^n\}\big |D^n, x\n1=x, s\n1=1\big].$$
\end{repeatlemma}

\proof{Proof} 
By Lemma \ref{lem:KG_add_invariant}, we may set $s=1$ as the target of optimization.  
For all sampled points $i=1,...,n$, we have that $\st^n(x^i,1;x,s)=0$ and $\mu^n(x^i,1)=y^i$
therefore $\max \mu^n(\tXn)=\max Y^n$. Define $\bar Y^n = \max Y^n$.
The expression for Knowledge Gradient becomes
\begin{eqnarray*}
    \CRNKG_n(x,s; X^n) &=& \E\big[\max\{ \bar Y^n, \mu^n(x,1) + \st^n(x,1;x,s)Z \}\big] - \max\{ \bar Y^n, \mu^n(x,1)\} \\
    &=& \E\big[\max\{ 0, \mu^n(x,1) + \st^n(x,1;x,s)Z -\bar Y^n \}\big] \\
    &=& \Delta(x)\Phi\left(\frac{\Delta(x)}{|\st^n(x,1;x,s)|}\right) - |\st^n(x,1;x,s)|\phi\left(\frac{\Delta(x)}{|\st^n(x,1;x,s)|}\right) \\
    &=& f\big(\Delta(x),\,\, |\st^n(x,1;x,s)|\big)
\end{eqnarray*}
where $\Phi(\cdot),\phi(\cdot)$ are cumulative and density functions
of the Gaussian distribution, $\Delta(x) = \mu^n(x,1) - \bar Y^n$ and
$f(a,b)$ is the well known expected improvement acquisition function
derived from the expectation of a truncated Gaussian random variable.
Note that the function $f(a,b)$ is monotonically increasing in $b$,
$ \frac{d}{db}f(a,b) = \phi(-a/b)>0.$
Hence, to prove the lemma, it is sufficient to show 
$|\st^n(x,1;x,1)| > |\st^n(x,1;x,2)|$
for all $x\in X$.
Firstly we may simplify $\st^n(x,1;x,1)$ as follows
\begin{eqnarray}
    \st^n(x,1;x,1) &=& k^n(x,1,x,1) / \sqrt{ k^n(x,1,x,1)} \\
    &=& \sqrt{k^n(x,1, x,1)}.
\end{eqnarray}
Substituting this into the inequality yields
\begin{eqnarray*}
    |\st^n(x,1;x,1)| &>& |\st^n(x,1;x,2)| \\
    \sqrt{k^n(x,1,x,1)} &>& \frac{|k^n(x,1,x,2)|}{\sqrt{k^n(x,2,x,2)}} \\
    1 &>& \frac{|k^n(x,1,x,2)|}{\sqrt{k^n(x,2,x,2)k^n(x,1,x,1)}} \\
    -1&<&\text{corr}(\theta(x,1),\theta(x,2)|D^n) \leq 1
\end{eqnarray*}
where the last line is true by the positive semi-definiteness of the 
kernel, the correlation between two random variables cannot be greater than one.
The above result demonstrates that allocating samples according to $\CRNKG$ will always sample
seed $s=1$. The target is stochastic however the objective is deterministic and the new output 
$y\n1\sim N(\mu^n(x,1), k^n(x,1,x,1))$. The acquisition function simplifies to
\begin{eqnarray*}
    \CRNKG_n(x,1; X^n) &=& \E\big[\max\{ 0, \mu^n(x,1) + \sqrt{k^n(x,1,x,1)}Z -\bar Y^n\}\big] \\
    &=& \E\big[\max\{ 0, y\n1 - \bar Y^n\}\big |D^n, x\n1=x, s\n1=1\big]
\end{eqnarray*}
where the last line is exactly the EI acquisition
criterion of \cite{jones1998efficient}.
\Halmos
\endproof
%
%
%

%
%
%
%
%
%

\clearpage
\section{Further Experimental Results}

\def\w{0.70}
\begin{figure}[h!]
    \centering
    \includegraphics[width=\w\textwidth]{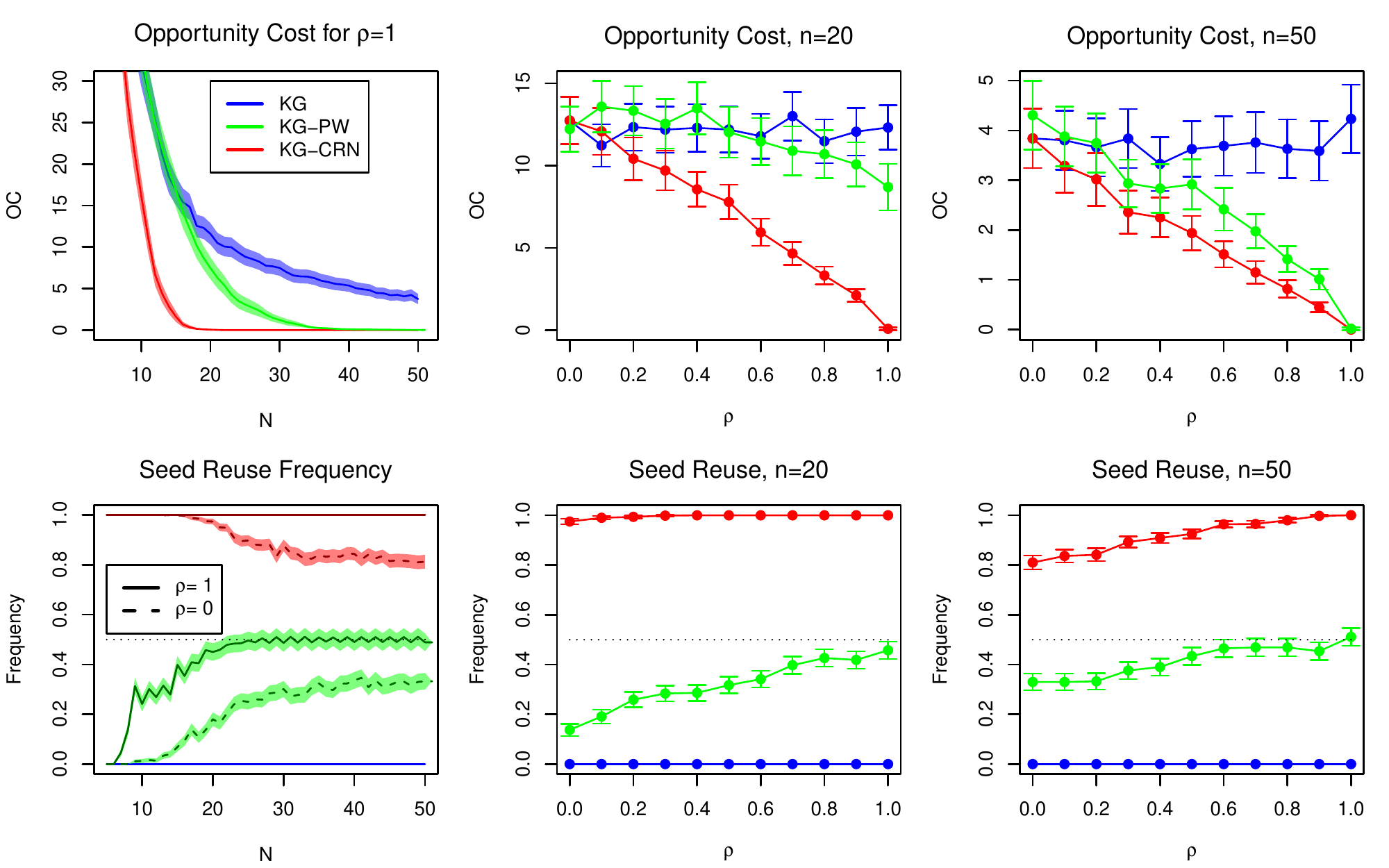}
    \caption{\label{fig:EC_offset_experiments}GP synthetic data with offsets and white noise ($\eta,\sigma_w\geq 0, \sigma_b=0$) only $\rho=\eta^2/(\eta^2+\sigma_w^2)$
    holding $\eta^2+\sigma_w^2=50^2$ constant. For low $\rho$, all algorithms perform similarly. As $\rho$
    increases, $\CRNKG$ samples more old seeds and outperforms other methods, $\PWKG$ samples singles first, duplicating
    KG and sampling doubles later improving upon KG.}
    
\end{figure}

\begin{figure}[h!]
    \centering
    \includegraphics[width=\w\textwidth]{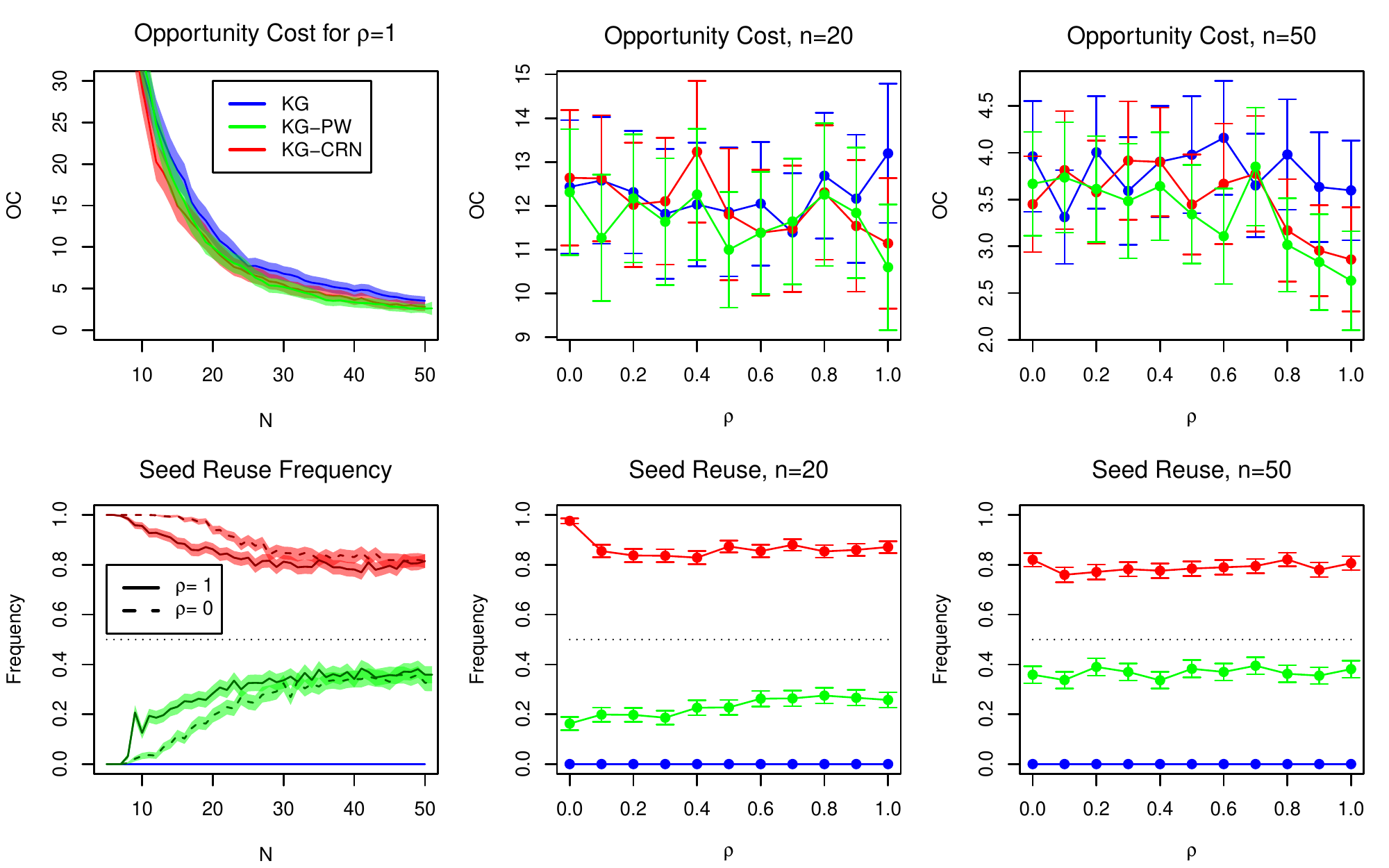}
    \caption{\label{fig:EC_wiggle_experiments}
    GP synthetic data generated with $\eta^2=0$ and $\rho=\sigma_b^2/(\sigma_b^2+\sigma_w^2)$
    holding $\sigma_b^2+\sigma_w^2=50^2$ constant. There is no significant benefit from bias functions alone in case of no offsets.
    }
\end{figure}

\def\ww{0.83}
\begin{figure}[h!]
    \centering
    \includegraphics[width=\ww\textwidth]{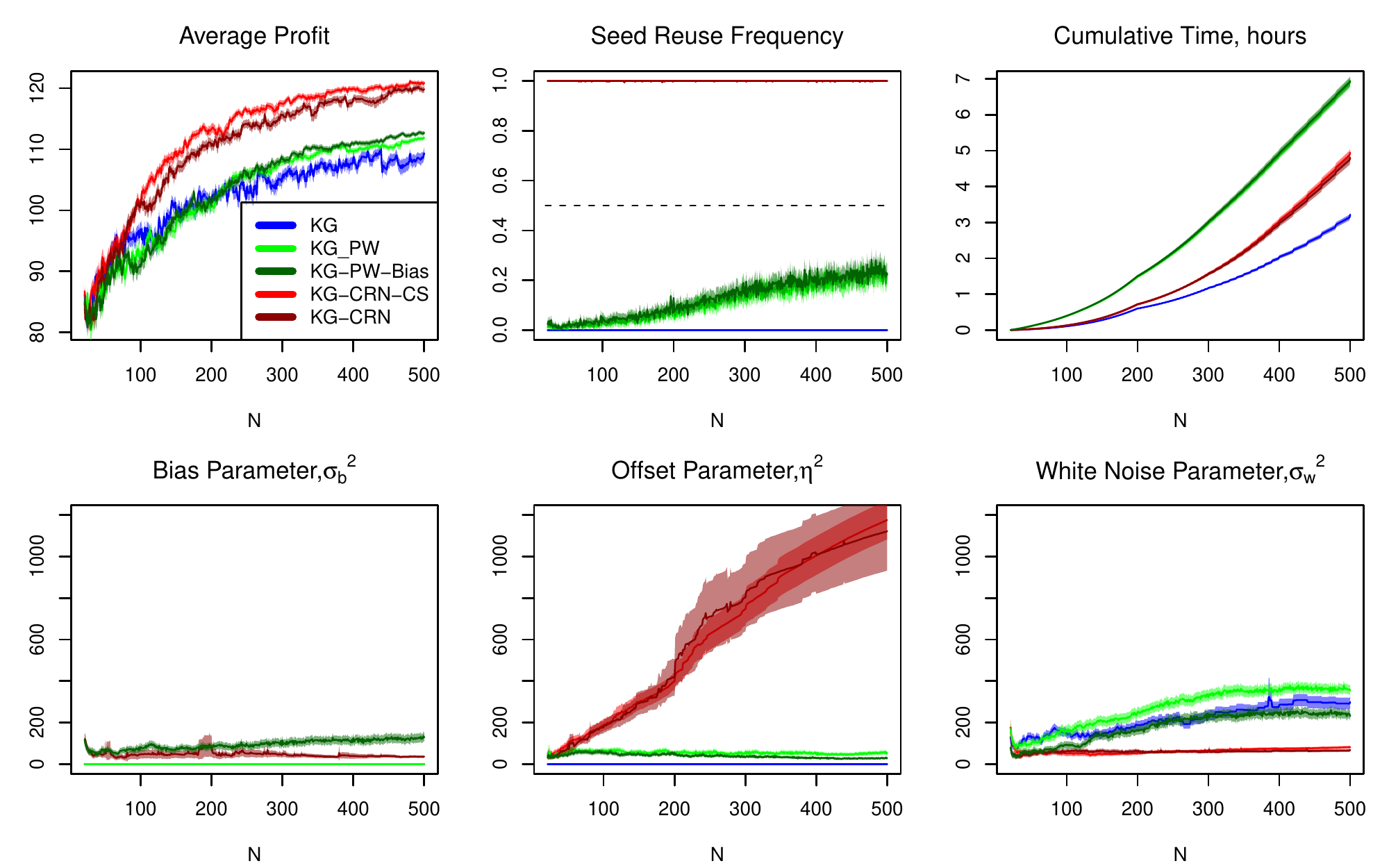}
    \caption{
    \label{fig:EC_ATO_diagnostics}
    ATO results. $\PWKG$ samples singles early on, and never
    learns a large offset parameter $\eta^2$. $\CRNKG$ samples
    old seeds and eventually learns a large offset parameter and never samples any new seeds. 
    KG has smallest runtime, followed by $\CRNKG$ variants then $\PWKG$ variants.
    }
\end{figure}

\begin{figure}[h!]
    \centering
    \includegraphics[width=\ww\textwidth]{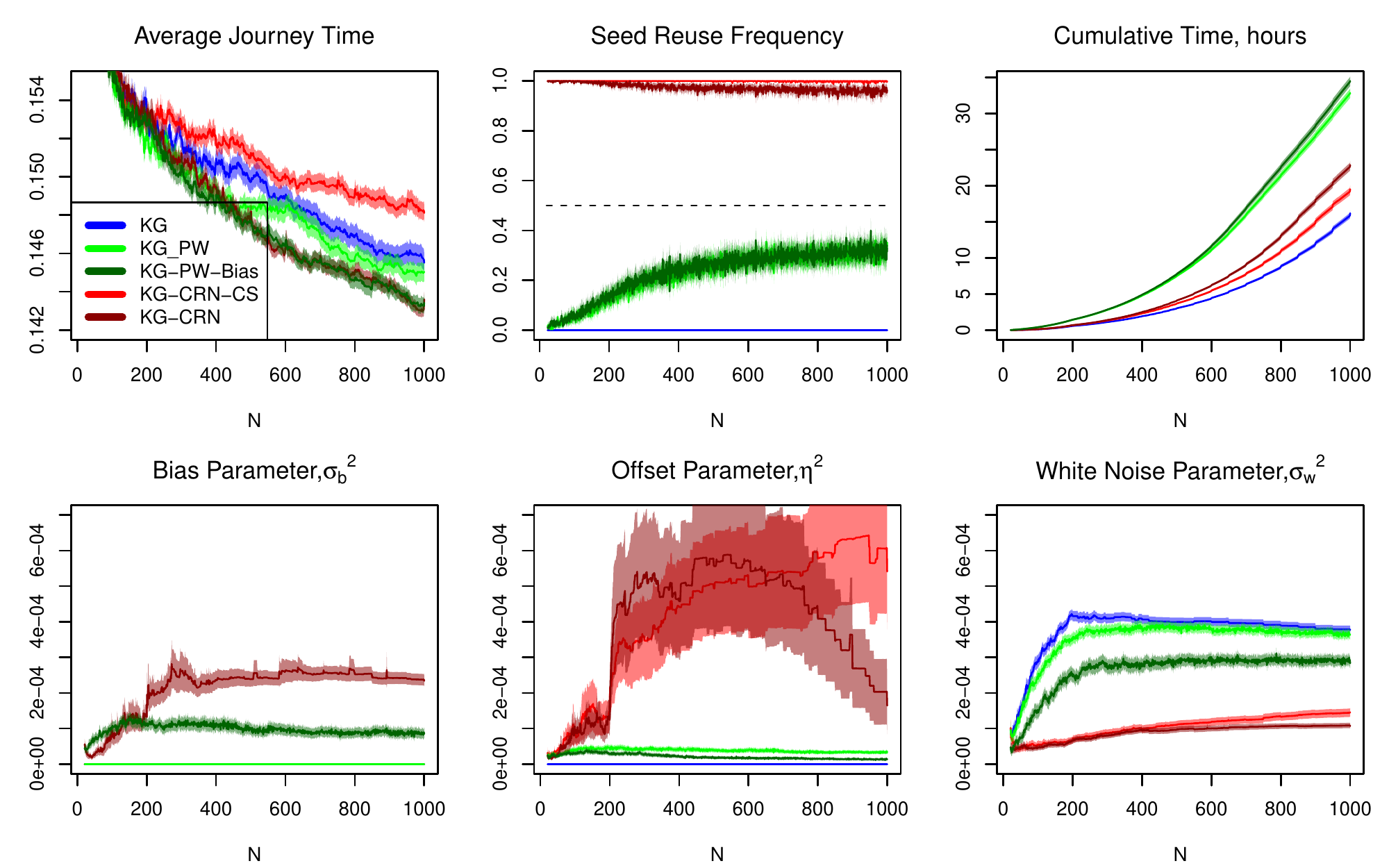}
    \caption{
    \label{fig:EC_damb_diagnostics}
    Ambulances in a square problem (AIS). The bias functions provide significant benefit to both $\CRNKG$ and $\PWKG$. Excluding bias functions, $\CRNKG-CS$,
    leads to inefficiently sampling only old seeds. The $\CRNKG$ variants learn larger offset parameters and require less computation time.
    }
\end{figure}

\begin{figure}[h!]
    \centering
    \includegraphics[width=\ww\textwidth]{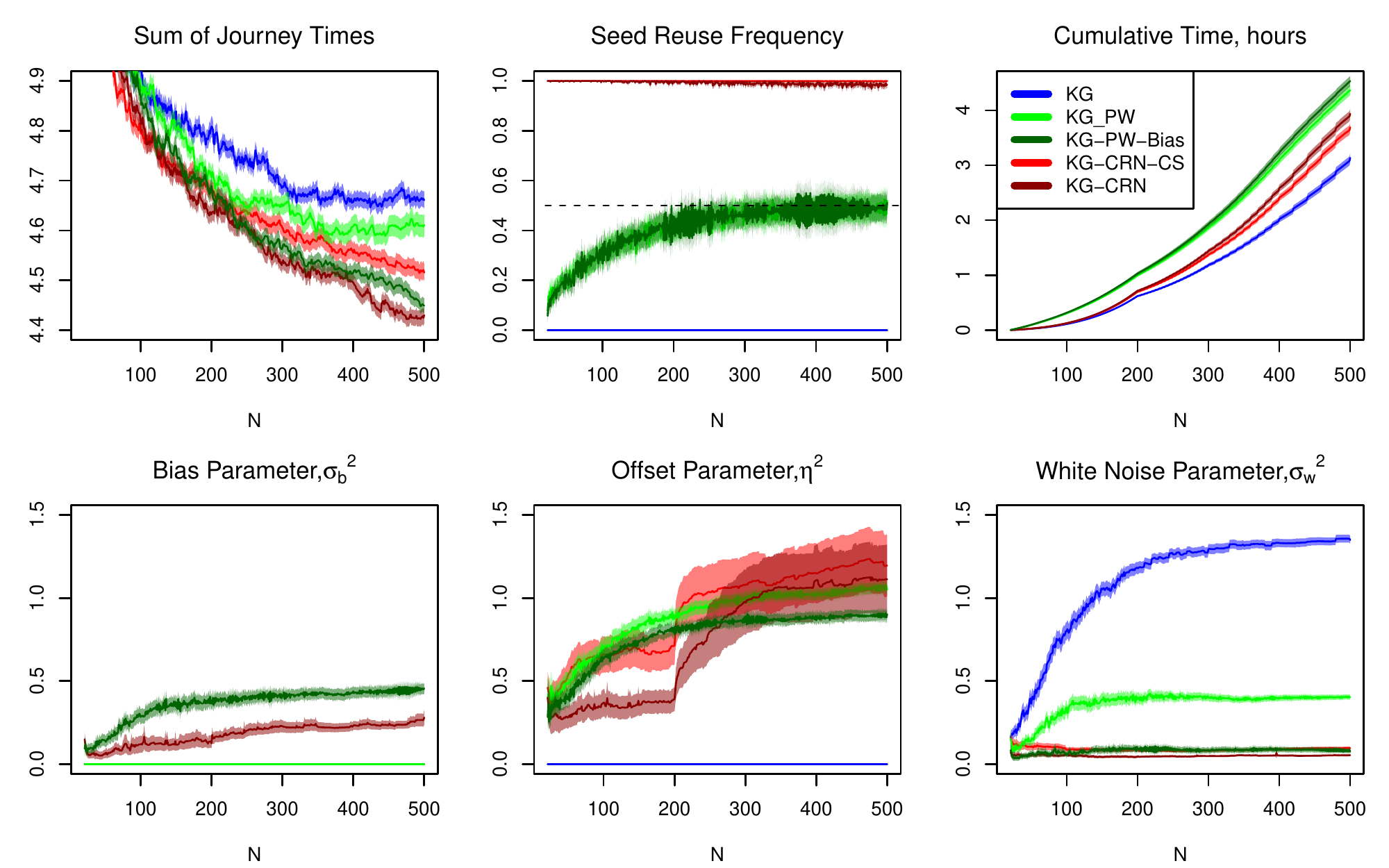}
    \caption{
    \label{fig:EC_amb_sum_delays}
    The AIS problem with the sum of journey times in a simulation as the objective. $\CRNKG$ variants improve performance over KG, and bias functions improve performance over compound spheric variants. Seed reuse is almost maximized by all methods.
    }
\end{figure}

\begin{figure}[h!]
    \centering
    \includegraphics[width=\ww\textwidth]{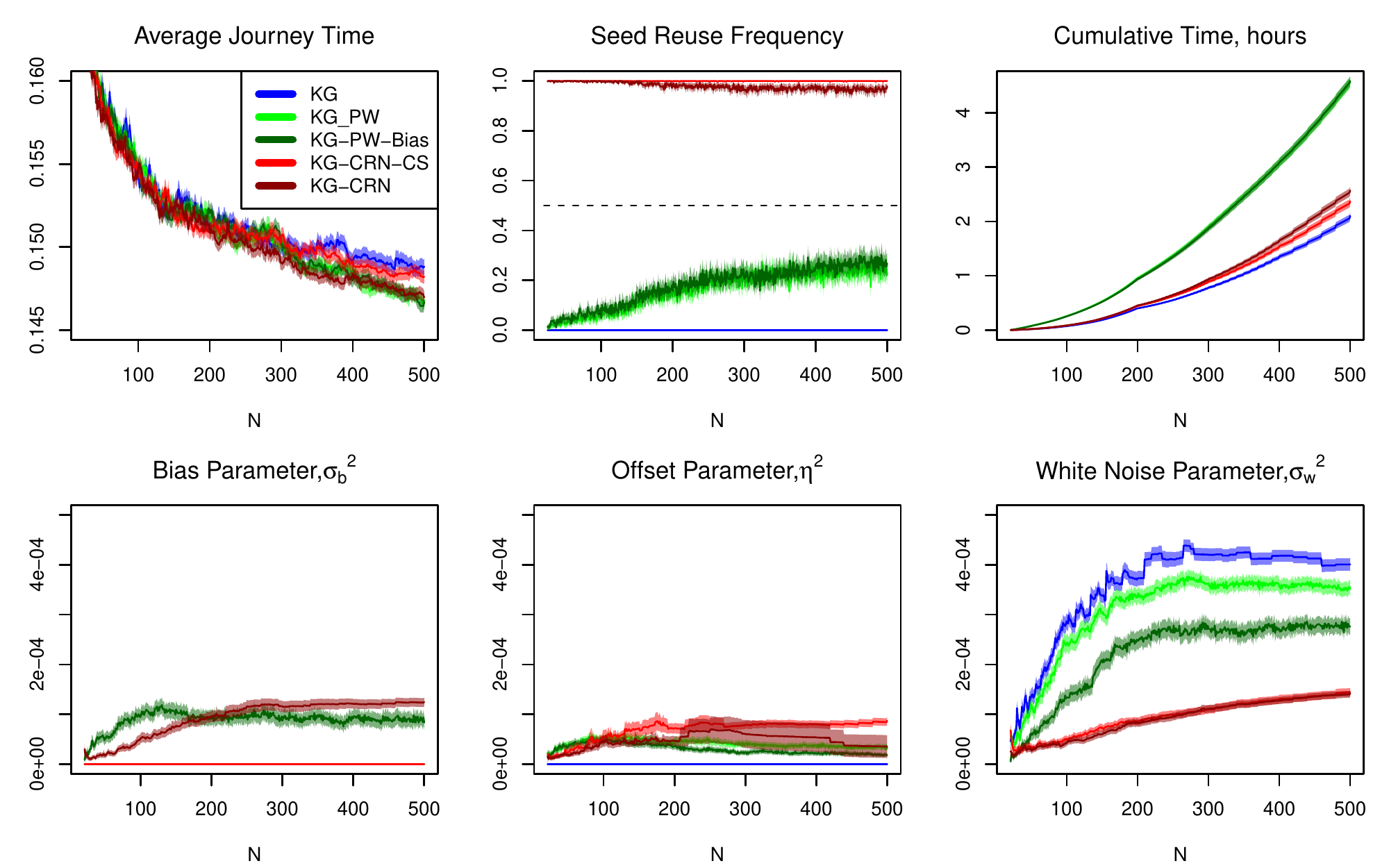}
    \caption{
    \label{fig:EC_amb_fixed_patients}
    All algorithm variants perform similarly and the offset and bias parameters are much lower than the white noise parameter suggesting there is little exploitable structure in the noise for this problem.
    }
\end{figure}

\clearpage

\section{Algorithm Implementation Details}
\subsection{Hyperparameter Learning}
The hyperparameters of the GP prior are estimated by multi-start conjugate gradient ascent of the marginal likelihood \citep{rasmussen2003gaussian}. 1000 random points are randomly uniformly distributed in a box
bounded below by zero in all dimensions above by double the bounding box side length for length scales, and $1.5(\max Y^n - \min Y^n)^2$ for all other parameters. The best 20 points are used for 100 steps of conjugate gradient ascent. This expensive search is used for all iterations up to 200, then at decreasing intervals thereafter to save computation time. iterations that are not in schedule, 20 steps of gradient ascent is applied using the current best hyperparameters as a starting point.
\begin{eqnarray*}
\P[Y^n|\tilde X^n, L, \sigma_{\tb}^2, \eta^2, \sigma_b^2, \sigma_w^2] 
&=& -\frac{1}{2}\left(   (Y^n-\bar Y)\trans K^{-1}(Y^n-\bar Y) + \log(|K|) + n\log(2\pi) \right) \\
K_{ij} &=& \sigma_{\tb}^2\exp\left(-\frac{1}{2}(x^i-x^j)\trans L(x^i-x^j)\right) \\
&& + \One_{s^i=s^j}\left( \eta^2 + \sigma_b^2\exp\left(-\frac{1}{2}(x^i-x^j)\trans L(x^i-x^j)\right) + \One_{x^i=x^j}\sigma_w^2\right).
\end{eqnarray*}

Firstly, an independent noise model (IND) is fitted by clamping
$\eta^2= \sigma_b^2=0$ to yield
\begin{equation}
{L^{IND}, \sigma^2_{\tb}}^{IND}, {\sigma_w^{2}}^{IND} = \amax{}\P[Y^n|\tilde X^n,  L, \sigma_{\tb}^2, \eta^2= \sigma_b^2=0, \sigma_w^2].
\end{equation}

Secondly, the noise parameters $\eta^2, \sigma_b^2, \sigma_w^2$ are
optimized whilst keeping  the total noise fixed 
$\eta^2+ \sigma_b^2+ \sigma_w^2={\sigma_w^2}^{IND}$ which is a 
two-dimensional optimization, we reparameterize as follows
\begin{eqnarray*}
\eta^2(\alpha, \beta)     &=& \beta(1-\alpha){\sigma_w^{2}}^{IND}\\
\sigma_b^2(\alpha, \beta) &=& (1-\beta)(1-\alpha){\sigma_w^2}^{IND}\\
\sigma_w^2(\beta) &=& \alpha{\sigma_w^2}^{IND} \\
\alpha, \beta &=& \amax{[0,1]^2}\P[Y^n|\tilde X^n, L^{IND}, {\sigma^2_{\tb}}^{IND}, 
\eta^2(\alpha, \beta), \sigma_b^2(\alpha, \beta), \sigma_w^2(\beta)]
\end{eqnarray*}

Thirdly, the final estimates of all hyperparameters are simultaneously fine-tuned
by gradient ascent. This three-stage method
guarantees that the found likelihood is greater than the
equivalent non-CRN parameter estimates. Note that the second
extra step of optimization is performed only over the unit square
and is thus cheaper than learning all hyperparameters from scratch.
\subsection{Optimization of $\CRNKG_n(x,s)$}\label{EC:hp_optim}
Derivatives of $\CRNKG$ and $\PWKG$, when evaluated by discretization
over $X$ as we do, are easily (but tediously) derived and can be found
in multiple previous works \citep{KGCP_scott2011correlated,xie2016bayesian}.
Alternatively, any automatic differentiation package, (Autograd, 
TensorFlow, PyTorch) may be used as the
mathematical operations are all common functions.
We propose the following optimization procedure:
\begin{enumerate}
    \item Evaluate $\CRNKG(x,s)$ across an initial Latin
Hypercube design with 1000 points over the acquisition space 
$\tilde X_{acq} = X\times \{1,...,\max S^n +1\}$.

\item Use the top 20 initial points to initialize 100 steps of
conjugate gradient ascent over $X$, holding the seed constant within each
run.
\item For the largest $(x,s)$ pair found, evaluate $\CRNKG(x,s)$ for the same $x$ on all seeds $s\in \{1,...,\max S^n +1\}$
\item Perform 20 steps of gradient ascent to fine
tune the $x$ from the best seed.
\end{enumerate}
When not using common random
numbers, stages one and two  use the same new seed and stages 
three and four are omitted.


\end{document}